\documentclass[10pt,twocolumn,letterpaper]{article}

\usepackage{cvpr}              % To produce the CAMERA-READY version
\usepackage{graphicx}
\usepackage{amsmath}
\usepackage{amssymb}
\usepackage{booktabs}
\usepackage{multirow}
\usepackage{makecell}
\usepackage{color}
\usepackage{subfloat}
\usepackage{bbding}
\usepackage{soul}
\usepackage[accsupp]{axessibility}  % Improves PDF readability for those with disabilities.
\usepackage[pagebackref,breaklinks,colorlinks]{hyperref}

\definecolor{MyGreen}{RGB}{30, 232, 114}
\definecolor{MyBlue}{RGB}{16, 83, 227}
\definecolor{MyRed}{RGB}{227, 16, 55}

\usepackage[capitalize]{cleveref}
\crefname{section}{Sec.}{Secs.}
\Crefname{section}{Section}{Sections}
\Crefname{table}{Table}{Tables}
\crefname{table}{Tab.}{Tabs.}

\def\cvprPaperID{320} % *** Enter the CVPR Paper ID here
\def\confName{CVPR}
\def\confYear{2023}

\begin{document}

\title{DBARF: Deep Bundle-Adjusting Generalizable Neural Radiance Fields}

% \author{Yu Chen\\
% National University of Singapore\\
% {\tt\small chenyu@comp.nus.edu.sg}
% \and
% Gim Hee Lee\\
% National University of Singapore\\
% {\tt\small gimhee.lee@comp.nus.edu.sg}
% }

\author{
Yu Chen \qquad Gim Hee Lee\\
Department of Computer Science, National University of Singapore\\
{\tt\small \{chenyu, gimhee.lee\}@nus.edu.sg}
}

\maketitle

%%%%%%%%% ABSTRACT
\begin{abstract}
Recent works such as BARF and GARF can bundle adjust camera poses with neural radiance fields (NeRF) which is based on coordinate-MLPs. 
Despite the impressive results, these methods cannot be applied to Generalizable NeRFs (GeNeRFs)  which require image feature extractions that are often based on more complicated 3D CNN or transformer architectures.
In this work, we first analyze the difficulties of jointly optimizing camera poses with GeNeRFs, and then further propose our DBARF to tackle these issues. Our DBARF which bundle adjusts camera poses by taking a cost feature map as an implicit cost function can be jointly trained with GeNeRFs in a self-supervised manner. Unlike BARF and its follow-up works, which can only be applied to per-scene optimized NeRFs and need accurate initial camera poses with the exception of forward-facing scenes, our method can generalize across scenes and does not require any good initialization. Experiments show the effectiveness and generalization ability of our DBARF when evaluated on real-world datasets. Our code is available at \url{https://aibluefisher.github.io/dbarf}. %the %\href{https://aibluefisher.github.io/dbarf}{project website}. %\footnote{\url{https://aibluefisher.github.io/dbarf}}.

\end{abstract}

%%%%%%%%% BODY TEXT
\section{Introduction}
\label{sec:intro}

The recent introduction of NeRF (Neural Radiance Fields)~\cite{DBLP:conf/eccv/MildenhallSTBRN20} bridges the gap between computer vision and computer graphics with the focus on the Novel view synthesis (NVS) task. 
NeRF demonstrates impressive capability of encoding the implicit scene representation and rendering high-quality images at novel views with only a small set of coordinate-based MLPs. Although NeRF and its variants simplify the dense 3D reconstruction part of the traditional photogrammetry pipeline that includes: the reconstruction of dense point clouds from posed images followed by the recovery and texture mapping of the surfaces into just a simple neural network inference, they still require known accurate camera poses as inputs.

\begin{figure}[htbp]
    \centering

    \includegraphics[width=1.0\linewidth]{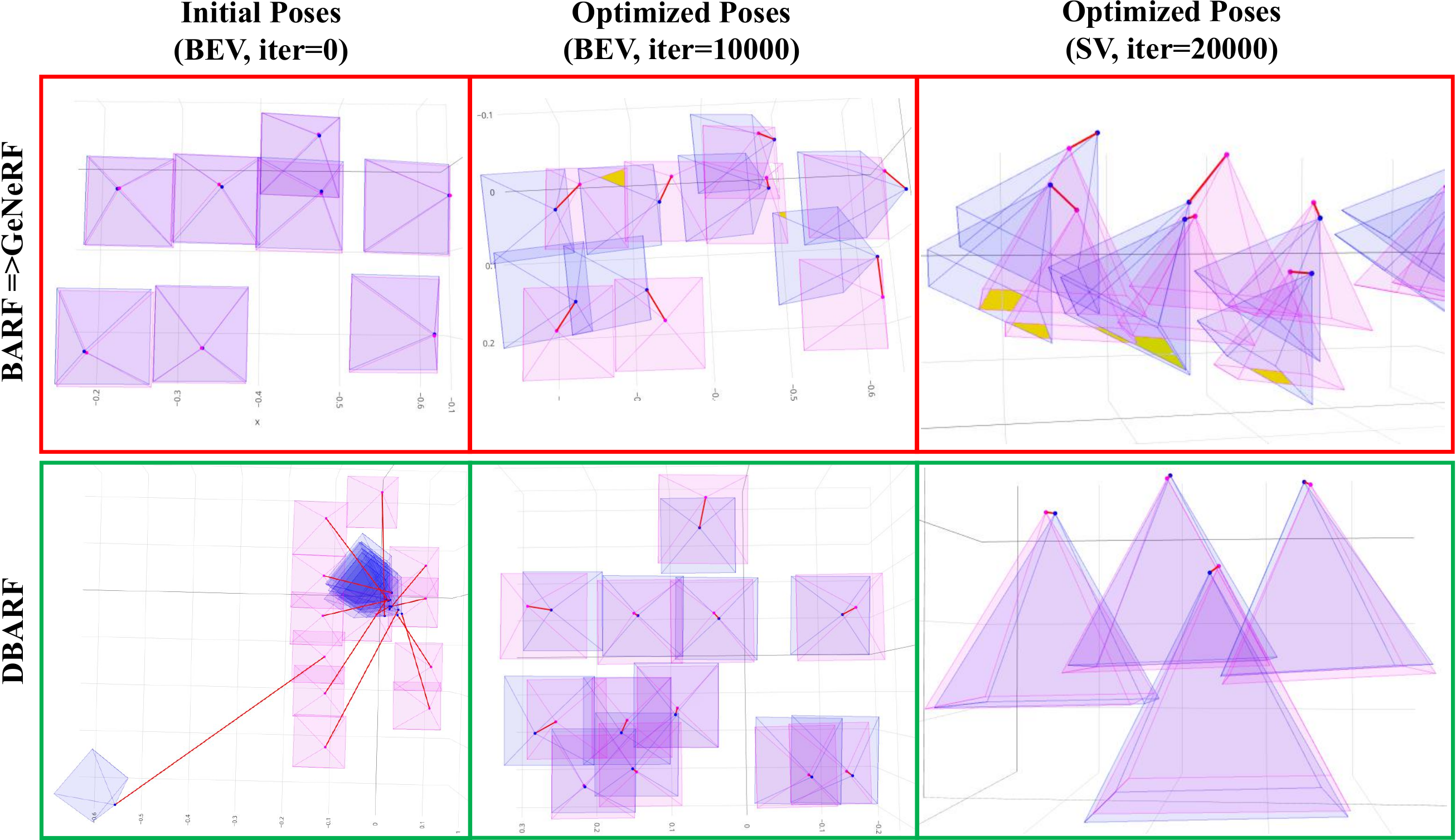}
    \vspace{-2mm}
    \caption{\textbf{Results of optimizing camera poses with BARF and DBARF}.
             From left to right are the initial camera poses, bird's eye view (BEV) of optimized camera poses after 
             $1e4$ iterations, and side view (SV) of optimized camera pose after $2e4$ iterations.
             Red and blue denote ground truths and estimated camera poses (The inconsistent ground truth poses in different iterations are due to the randomness of selecting the training batches).
             \textbf{Top}: The camera poses diverge quickly when BARF~\cite{DBLP:conf/iccv/LinM0L21} is applied to GeNeRF, even with the camera poses initialized by perturbing the ground truth with very small noise.
             \textbf{Bottom}: Results obtained by our DBARF, the camera poses are randomly initialized.
             }
             \vspace{-6mm}
    \label{fig:teaser}
\end{figure}

Nonetheless, the acquisition of camera poses is expensive in the real world. Most NeRF-related methods obtain the camera poses by Structure-from-Motion (SfM)~\cite{DBLP:conf/cvpr/SchonbergerF16,DBLP:conf/iccv/LindenbergerSLP21,DBLP:journals/pr/ChenSCW20}.
In SfM, camera poses are optimized under the keypoint-metric reprojection error in a process referred to as bundle adjustment~\cite{DBLP:conf/iccvw/TriggsMHF99}. A notorious problem of SfM is that it sometimes fails, \eg in textureless or self-similar scenes, and can also take days or even weeks to complete for large-scale scenes. Consequently, one main forthcoming issue with NeRF is that its rendering quality highly relies on accurate camera poses. Recently, several works try to solve the pose inaccuracy jointly with NeRF. One of the representative works is BARF~\cite{DBLP:conf/iccv/LinM0L21}. 
NeRF maps the pixel coordinates into high-dimensional space as Fourier features~\cite{DBLP:conf/nips/TancikSMFRSRBN20} before inputting into the MLPs to enable networks to learn the high-frequency part of images. However, Fourier features can be a double-edged sword when the camera poses are jointly optimized with NeRF, where gradients from high-frequency components dominate the low-frequency parts during training. To mitigate this problem, BARF draws inspiration from the non-smoothness 
optimization in high-dimensional functions: optimizer can get stuck at a local optimum, but the training can be easier when the objective function is made smoother.
Consequently, BARF adopts a coarse-to-fine strategy which first masks out the high-frequency components, and then gradually reactivates them after the low-frequency components become stable. The camera poses are adjusted by the photometric loss during training instead of the keypoint-metric cost in SfM. Despite its promising results, BARF and its follow-up works~\cite{DBLP:conf/iccv/MengCLW0X0Y21,DBLP:journals/corr/abs-2204-05735} still require the pre-computed camera poses from SfM.

One other issue with vanilla NeRF is that it needs time-consuming per-scene training. Making NeRF generalizable across scenes~\cite{DBLP:conf/cvpr/YuYTK21,DBLP:conf/cvpr/WangWGSZBMSF21,DBLP:conf/iccv/ChenXZZXY021,DBLP:conf/cvpr/JohariLF22} has recently gained increasing attention. However, similar to vanilla NeRF, GeNeRFs (generalizable NeRFs) also depend on accurate camera poses. There is no existing work that tried to optimize the camera poses jointly with GeNeRFs. \textbf{This intrigues us to investigate the replacement of NeRF with GeNeRFs in BARF.} We find that the joint optimization is non-trivial in our task settings, and the camera poses can diverge quickly even when initialized with the ground truths  (\cf top row of Fig.~\ref{fig:teaser}). 

In this paper, we identified two potential reasons which cause the failure of bundle adjusting GeNeRFs. The first reason is the aggregated feature outliers, which are caused by occlusions. The other reason is due to the high non-convexity of the cost function produced by ResNet features~\cite{DBLP:conf/iclr/TangT19}, which produces incoherent displacements like the issue caused by positional encodings~\cite{DBLP:conf/nips/TancikSMFRSRBN20} in BARF. We further proposed our method DBARF, which jointly optimizes GeNeRF and relative camera poses by a deep neural network. Our implicit training objective can be equivalently deemed as a smooth function of the coarse-to-fine training objective in BARF. Specifically, we construct a residual feature map by warping 3D points onto the feature maps of the nearby views. We then take the residual feature map as an implicit cost function, which we refer to as \textit{cost map} in the following sections. By taking the cost map as input, we utilize a deep pose optimizer to learn to correct the relative camera poses from the target view to nearby views. We further jointly train the pose optimizer and a GeNeRF with images as supervision, which does not rely on ground truth camera poses. In contrast to previous methods which only focus on per-scene camera pose optimization, our network is generalizable across scenes.

In summary, the contributions of this work are:
\begin{itemize}
    \item We conduct an experiment on bundle adjusting GeNeRFs by gradient descent and analyze the difficulty of jointly optimizing camera poses with GeNeRFs.
    \item We present DBARF to deep bundle adjusting camera poses with GeNeRFs. The approach is trained end-to-end without requiring known absolute camera poses.
    \item We conduct experiments to show the generalization ability of our DBARF, which can outperform BARF and GARF even without per-scene fine-tuning.
\end{itemize}

\begin{figure*}[htbp]
    \centering
    \subfloat[] 
    {
        \includegraphics[width=0.4\linewidth]{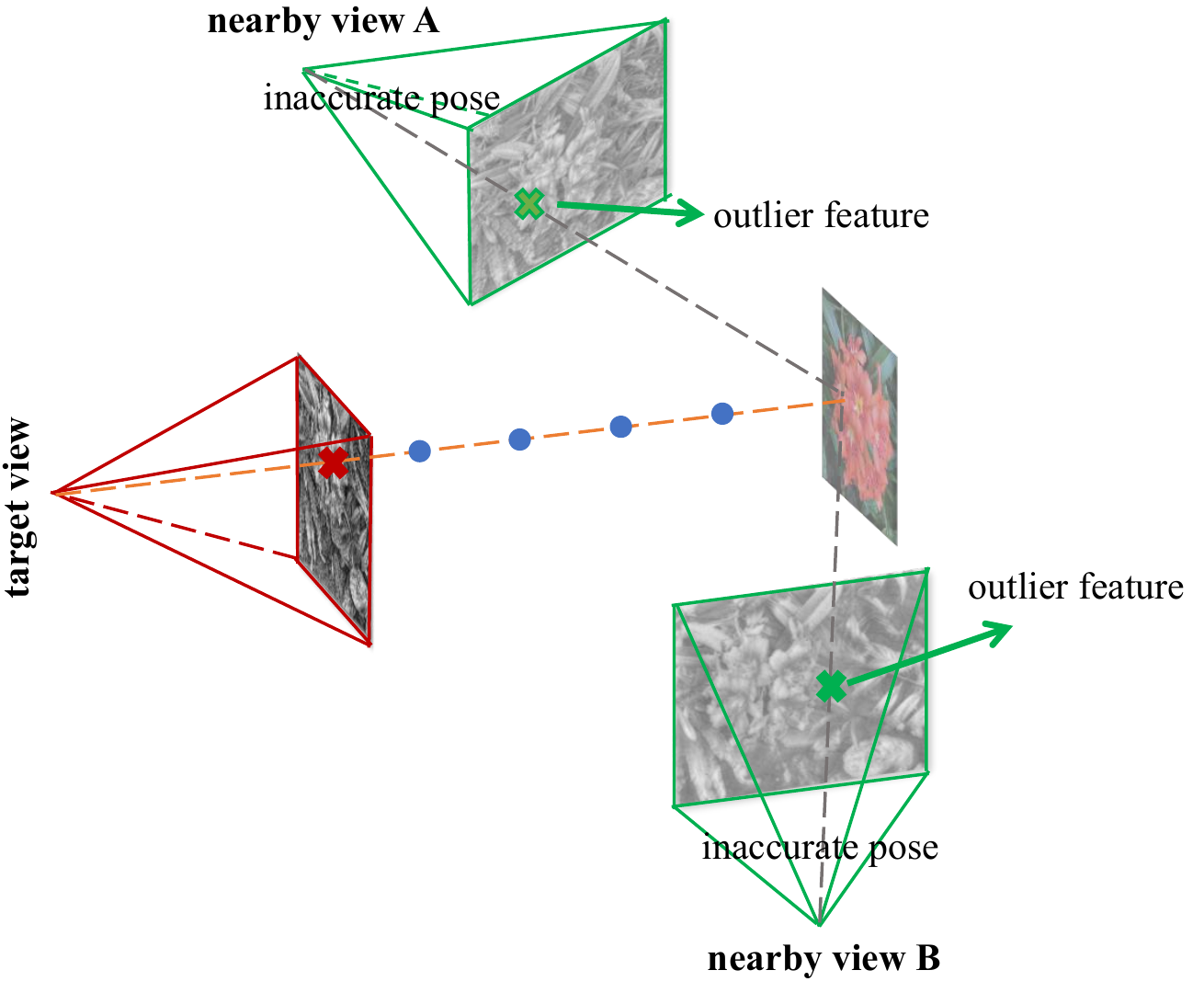}
        \label{fig:generf_outlier_feature}
    }
    \subfloat[] 
    {
        \includegraphics[width=0.18\linewidth,height=0.36\textwidth]{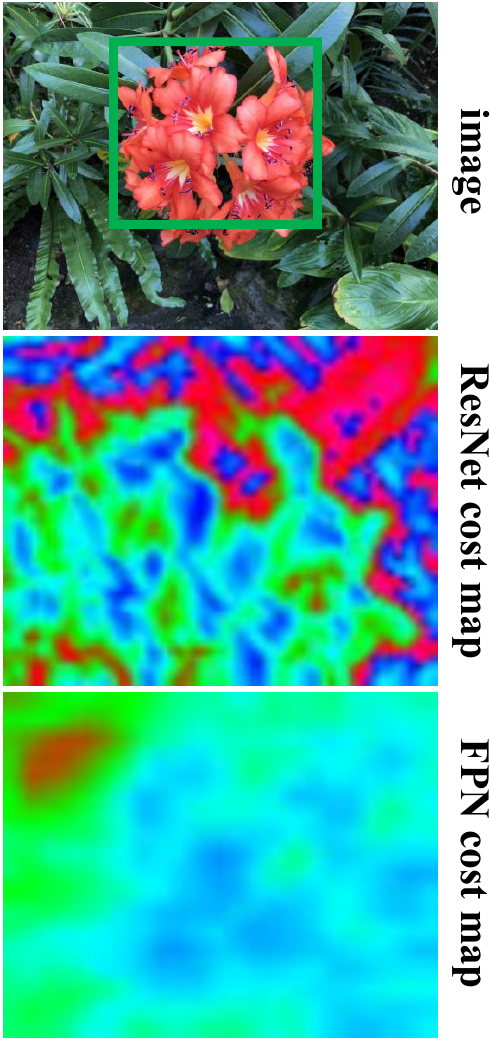}
        \label{fig:reset_feature_vs_fpn_feature}
    }
    \subfloat[]
    {
        \includegraphics[width=0.4\linewidth]{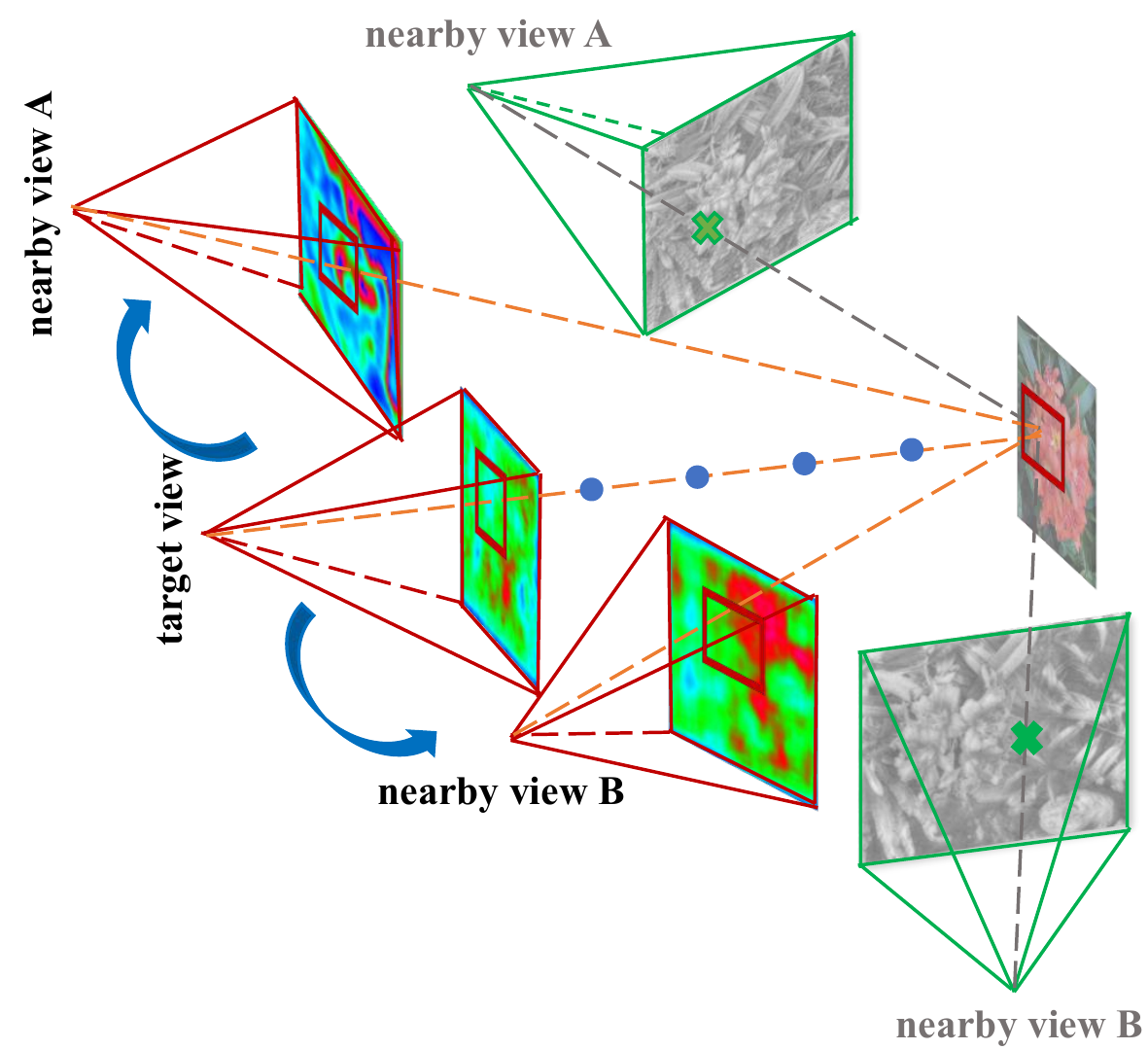}
        \label{fig:deep_optimizer_patch}
    }
    \vspace{-2mm}
    \caption{\textbf{The difficulties when optimizing camera poses with GeNeRFs}:
             a) Sampled features tend to be outliers when they are occluded.
             b) ResNet gives a non-smooth cost feature map (middle) while feature patches with FPN lead to a smoother cost map (bottom)
             c) Our method sample image patches to predict relative camera poses.}
    \vspace{-4mm}
    \label{fig:difficulty_on_optimizing_camera_poses}
\end{figure*}

\section{Related Work}
\label{sec:related_work}

\paragraph{Novel View Synthesis.} Given posed images, vanilla NeRF~\cite{DBLP:conf/eccv/MildenhallSTBRN20} used an MLP to predict the volume density and pixel color for a point sampled at 3D space. The low-dimensional inputs (point coordinates and ray directions) are encoded by the positional encodings~\cite{DBLP:conf/nips/TancikSMFRSRBN20} to high-dimensional representations, such that the network can learn high-frequency components of images. 
While NeRF~\cite{DBLP:conf/eccv/MildenhallSTBRN20} and later follow-up works achieved great progress in improving the rendering quality, 
such as the anti-aliasing effects~\cite{DBLP:journals/corr/abs-2010-07492,DBLP:conf/eccv/MildenhallSTBRN20,DBLP:conf/cvpr/BarronMVSH22} and 
reflectance~\cite{DBLP:conf/cvpr/VerbinHMZBS22}, reducing training time~\cite{DBLP:conf/iccv/ReiserPL021,DBLP:journals/tog/MullerESK22,
DBLP:conf/cvpr/Fridovich-KeilY22} and rendering time~\cite{DBLP:conf/nips/LiuGLCT20,DBLP:conf/iccv/YuLT0NK21,DBLP:conf/cvpr/0004SC22}, 
they still require time-consuming per-scene training.

Pixel-NeRF~\cite{DBLP:conf/cvpr/YuYTK21} is the first that generalizes NeRF to unseen scenes. It extracts image features from 
a feature volume by projection and interpolation, and then the image features are fed into a NeRF-like MLP network to obtain RGB color and 
density values. IBRNet~\cite{DBLP:conf/cvpr/WangWGSZBMSF21} aggregates per-point image feature from nearby views, 
the image features are weighted by a PointNet-like~\cite{DBLP:conf/cvpr/QiSMG17} architecture. Taking the weighted features as input, a ray transformer~\cite{DBLP:conf/nips/VaswaniSPUJGKP17} is further introduced to predict density, and another MLP is used to predict the pixel color. MVSNeRF~\cite{DBLP:conf/iccv/ChenXZZXY021} constructs 3D feature cost volume from $N$ depth hypothesis, then a neural voxel volume is reconstructed by a 3D CNN, pixel color and volume density are predicted by a MLP. 

GeoNeRF~\cite{DBLP:conf/cvpr/JohariLF22} extends MVSNeRF by using CasMVSNet~\cite{DBLP:conf/cvpr/GuFZDTT20} to let the network be aware of scene geometry. It adopts a similar approach as IBRNet~\cite{DBLP:conf/cvpr/WangWGSZBMSF21} to regress image color and volume density. NeuRay~\cite{DBLP:conf/cvpr/LiuPLWWTZW22} further predicts the visibility of 3D points to tackle the occlusion issue in previous GeNeRFs, and a consistency loss is also proposed to refine the visibility in per-scene fine-tuning. Instead of composing colors by volume rendering, LFNR~\cite{DBLP:conf/cvpr/SuhailESM22} and GPNR~\cite{DBLP:journals/corr/abs-2207-10662} 
adopts a 4D light field representation and a transformer-based architecture to predict the occlusions and colors for features aggregated from epipolar lines~\cite{DBLP:books/daglib/0015576}.

\vspace{-3mm}
\paragraph{Novel View Synthesis with Pose Refinement.}
I-NeRF~\cite{DBLP:conf/iros/LinFBRIL21} regressed single camera pose while requiring a pretrained NeRF model and matched keypoints 
as constraints. NeRF$--$~\cite{DBLP:journals/corr/abs-2102-07064} jointly optimizing the network of NeRF and camera pose embeddings, which 
achieved comparable accuracy with NeRF methods that require posed images. 
SiNeRF~\cite{DBLP:journals/corr/abs-2210-04553} adopts a 
SIREN-MLP~\cite{DBLP:conf/nips/SitzmannMBLW20} and a mixed region sampling strategy to circumvent the sub-optimality issue in NeRF$--$.
BARF~\cite{DBLP:conf/iccv/LinM0L21} proposed to jointly train NeRF with imperfect camera poses with a coarse-to-fine strategy. 
During training, the low-frequency components are learned at first and the high-frequency parts are gradually activated to alleviate gradient inconsistency issue. GARF~\cite{DBLP:journals/corr/abs-2204-05735} extends BARF with a positional-embedding less coordinate network. RM-NeRF~\cite{DBLP:journals/corr/abs-2210-04233} jointly trains a GNN-based motion averaging network~\cite{DBLP:conf/cvpr/Govindu04,DBLP:conf/eccv/PurkaitCR20} and Mip-NeRF
~\cite{DBLP:conf/iccv/BarronMTHMS21} to solve the camera pose refinement issue in multi-scale scenes. GNeRF~\cite{DBLP:conf/iccv/MengCLW0X0Y21} utilized an adversarial learning method to estimate camera poses. 
Camera poses in GNeRF are randomly sampled from prior-known camera distribution, and then a generator generates the corresponding fake images by volume rendering, together with a discriminator that classifies the real and fake images. An inversion network finally learns to predict the camera poses by taking the fake images as input. VMRF~\cite{DBLP:conf/mm/ZhangZWYZSZL22} can learn NeRF without known camera poses. The unbalanced optimal transport is introduced to learn the relative transformation between the real image and the rendered image, then camera poses are updated by the predicted relative poses to enable a finer training of NeRF.

None of the mentioned works can be applied to generalizable NeRFs and thus require time-consuming per-scene optimization. We also notice that there is a concurrent work~\cite{DBLP:journals/corr/abs-2210-07181} trying to make NeRF and pose regression generalizable. However, it only focuses on single-view rendering tasks. In contrast, we focus on the multiple views settings, which are more challenging than the single view.

%%%%%%%%%%%%%%%%%%%%%%%%%%%%%%%%%%%%%%%%%%%%%%%%%%%%%%%%%%%%% Notations %%%%%%%%%%%%%%%%%%%%%%%%%%%%%%%%%%%%%%%%%%%%%%%%%%%%%%%%%

\section{Notations and Preliminaries}
\label{sec:notations}
We follow the notations %convention 
in BARF~\cite{DBLP:conf/iccv/LinM0L21}. The image synthesis process is depicted by the equation below:
% \begin{small}
    \begin{equation}
        \hat{\mathbf{I}} = h \big(
                g(\omega(\mathbf{X}^1, \mathbf{P}); \mathbf{\Theta}),\ 
                % g(\omega(\mathbf{X}^2, \mathbf{P}); \mathbf{\Theta}),\ 
                \cdots,\ 
                g(\omega(\mathbf{X}^K, \mathbf{P}); \mathbf{\Theta})
        \big),
    \end{equation}
% \end{small}
where $\mathbf{X}^k = Z_k \mathbf{u}$ is a 3D point in the camera frame, $\{Z_1, Z_2, \cdots, Z_K\}$ are the sampled depths and $\mathbf{u}$ is the %homogeneous 
camera normalized pixel coordinates in the image. $h(\cdot)$ is the ray composition function, $g(\cdot)$ is the NeRF network, $\omega(\cdot)$ denotes the rigid transformation 
which projects the point $Z_k \mathbf{u}$ from the camera frame to the world frame by the camera pose $\mathbf{P}$,  $\mathbf{\Theta}$ denotes the network parameters. 

Once we obtained the point color $c_k$ and volume density $\sigma_k$ of all the $K$ points, the per-pixel RGB $C(\mathbf{r})$ and depth value $D(\mathbf{r})$ can be 
approximated with the quadrature rule:
% \begin{align}
\begin{subequations}
\begin{eqnarray}
    \label{equ:volume_rendering}
    & C (\mathbf{r}) = \sum_{k=1}^K T_k (1-\exp(-\sigma_k \delta_k)) c_k, \label{subequ:color_accumulation} \\
    & D (\mathbf{r}) = \sum_{k=1}^K T_k (1-\exp(-\sigma_k \delta_k)) Z_k, \label{subequ:depth_accumulation}%\\
    % & T_k = \exp(-\sum_{l=1}^{k-1} \sigma_l \delta_l),\ \delta_k = t_{k+1} - t_{k},
% \end{align}
\end{eqnarray}
\end{subequations}
where $T_k = \exp(-\sum_{l=1}^{k-1} \sigma_l \delta_l),\ \delta_k = Z_{k+1} - Z_{k}$
% \begin{equation}
%     T_k = \exp(-\sum_{l=1}^{k-1} \sigma_l \delta_l),\ \delta_k = Z_{k+1} - Z_{k}
% \end{equation}
%$T_{k}$ 
is the accumulated transmittance, and $\delta_k$ is the distance between adjacent samples. Please refer to~\cite{DBLP:conf/eccv/MildenhallSTBRN20} for more details on the volume rendering technique.

%%%%%%%%%%%%%%%%%%%%%%%%%%%%%%%%%%%%%%%%%%%%%%%%%%%%%%%%%% Method %%%%%%%%%%%%%%%%%%%%%%%%%%%%%%%%%%%%%%%%%%%%%%%%%%%%%%%%%%%%%%
\section{Our Method}
\label{sec:method}

\subsection{Generalizable Neural Radiance Field}

We adopt the term \textbf{GeNeRFs} to denote a bundle of Generalizable Neural Radiance Field methods~\cite{DBLP:conf/cvpr/YuYTK21,
DBLP:conf/cvpr/WangWGSZBMSF21,DBLP:conf/iccv/ChenXZZXY021,DBLP:conf/cvpr/JohariLF22}. Since these methods share a common philosophy in their
network architectures, we can abstract GeNeRFs into a series of high-dimensional functions.

GeNeRFs first extract 2D image features by projecting a point onto the feature map $\mathbf{F}_j$:
% \begin{small}
\begin{equation}
    \mathbf{f} = \chi \big( \Pi (\mathbf{P}_j, \omega(\mathbf{X}_i^k, \mathbf{P}_i)), \mathbf{F}_j \big),
\end{equation}
% \end{small}
where $\chi (\cdot)$ is the differentiable bilinear interpolation function, $\Pi(\cdot)$ is the reprojection function which 
maps points from world frame to image plane, $\mathbf{P}_i$ is the camera pose of image $i$, and $\mathbf{P}_j$ is the camera pose of image $j$ in the nearby view of image $i$. $\mathbf{X}_i^k=Z_{k} \mathbf{u}_i$ is the $k^{\text{th}}$ 3D point in image $i$, where $\mathbf{u}_i$ is the camera normalized pixel coordinates and $Z_{k}$ is depth of the $k^{th}$ 3D point in image $i$.

%Secondly, 
To render a novel view $i$, GeNeRFs either sample $K$ points and aggregate pixel-level features for each emitted ray, or 
construct a 3D cost volume by plane sweep~\cite{DBLP:conf/eccv/YaoLLFQ18,DBLP:conf/cvpr/GuFZDTT20}, from $M$ selected nearby views. 
%And then
Subsequently, per-depth volume density and pixel color are predicted by a neural network.
For clarity and without losing generality, we abstract the feature aggregation function $f_a(\cdot)$ as: %by:
\begin{equation}
    % \begin{matrix}
    % g_1 & = & f_a(\mathbf{f}_1^1, \mathbf{f}_2^1,\ \cdots,\ \mathbf{f}_M^1) \\ 
    % & \vdots & \\
    % g_K & = & f_a(\mathbf{f}_1^K, \mathbf{f}_2^K,\ \cdots,\ \mathbf{f}_M^K)
    % \end{matrix},
    g_k = f_a(\mathbf{f}_1^k, \mathbf{f}_2^k,\ \cdots,\ \mathbf{f}_M^k),
\end{equation}
% \begin{equation}
% % $$
% \left\{
% \begin{aligned}
%         & g_1 = f_a(\mathbf{f}_1^1, \mathbf{f}_2^1,\ \cdots,\ \mathbf{f}_M^1),\\
%         & \qquad\qquad \vdots \\
%         & g_K = f_a(\mathbf{f}_1^K, \mathbf{f}_2^K,\ \cdots,\ \mathbf{f}_M^K),
% \end{aligned}
% \right.
% % $$
% \end{equation}
where $\mathbf{f}_m^k$ denotes the feature vector of image point $\mathbf{u}$ sampled at depth $Z_k$ in image $m$ at the nearby view of image $i$.
The rendered target image is then given by:
% \begin{small}
\begin{equation}
    \hat{\mathbf{I}}_{\text{target}} := \hat{\mathbf{I}}_i = h(g_1,\, \cdots,\ g_K; \mathbf{\Phi}),
\end{equation}
% \end{small}
where $h(\cdot)$ is the GeNeRF network, and $\mathbf{\Phi}$ is the network parameters.

Similar to vanilla NeRF, the training loss for GeNeRFs is the photometric error between the rendered target image and the ground truth 
target image:
% \begin{small}
\begin{equation}
    \label{equ:photometric_loss}
    \mathcal{L}_{\text{rgb}} = \sum_i^N \sum_{\mathbf{u}}  \| \hat{\mathbf{I}}_i - \mathbf{I}_i (\mathbf{u}) \|.
\end{equation}
% \end{small}
$N$ is the total number of images in the training dataset.

\begin{figure*}[htbp]
    \centering
      \includegraphics[width=0.95\linewidth]{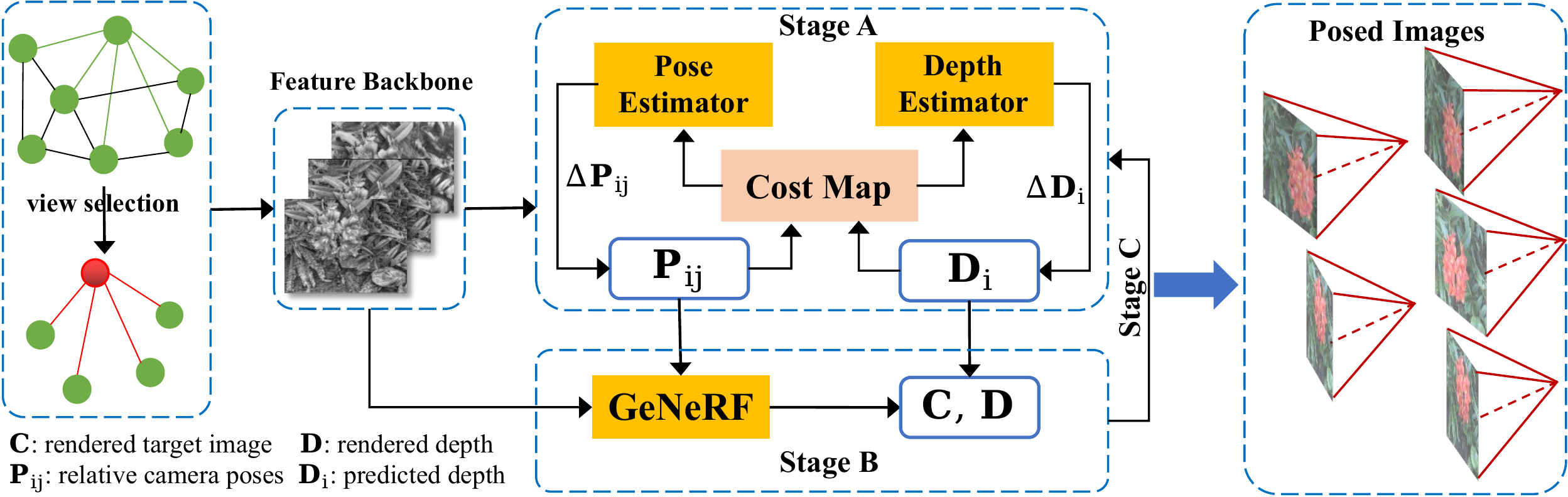}
      \vspace{-2mm}
    \caption{\textbf{Network architecture of our proposed DBARF}. The input is images and a scene graph. 1) Nearby views are selected from a scene graph since the camera poses are unknown. 2) Image features are extracted by ResNet-like~\cite{DBLP:conf/cvpr/HeZRS16} backbone. 3) In stage A, the image feature of the target view is warped to each nearby view by the corresponding current camera poses and depth, a cost map is constructed by the image feature difference. Camera poses and depth are recurrently optimized by taking the cost map as an implicit loss. 4) In stage B, we utilize a generalizable NeRF to predict image color and density value, and the final image is rendered by volume rendering. 5) In stage C, the pose optimizer and the generalizable NeRF are jointly learned. 6) Finally, our network outputs the posed images.}
    \vspace{-3mm}
    \label{fig:network_architecture}
  \end{figure*}

\subsection{Difficulties of Bundle Adjusting GeNeRFs}
\label{subsec:on_the_difficulty_of_ba_generfs}
BARF~\cite{DBLP:conf/iccv/LinM0L21} can jointly optimize NeRF with imperfect camera poses. 
The success of BARF can be largely attributed to the coarse-to-fine training strategy, which can deal with the 
gradient inconsistency between low-frequency components and high-frequency components. 
Specifically, the low-frequency components are first learned with the high-frequency part being masked out; then the high-frequency components are learned when the low-frequency components become stable. Otherwise, gradients from the high-frequency components, \ie high $k$'s tend to dominate the training process due to the positional encodings~\cite{DBLP:conf/nips/TancikSMFRSRBN20}:
\begin{equation}
    \frac{\partial \gamma_k (\mathbf{P})}{\partial \mathbf{P}} = 2^k \pi \cdot [-\sin (2^k \pi \mathbf{P}),\ \cos (2^k \pi \mathbf{P})],
    \label{equ:frequencyEncoding}
\end{equation}
where $\gamma_k (\mathbf{P}) = [\cos (2^k \pi \mathbf{P}),\ \sin (2^k \pi \mathbf{P})]$.

The fact that BARF and its variants~\cite{DBLP:journals/corr/abs-2102-07064,DBLP:journals/corr/abs-2210-04233,DBLP:journals/corr/abs-2210-04553} can optimize the camera poses by gradient descent jointly with NeRF intrigues us to ask the question: \textbf{Can we also directly optimize the camera poses jointly with GeNeRFs by gradient descent just like BARF?} To answer the question, we adopt a pretrained GeNeRF model and construct a $N \times 6$ learnable pose embedding like BARF. The pose embedding is jointly trained with the GeNeRF model and optimized by Adam with a learning rate of $1e-5$. Unfortunately, we found the camera 
poses drifted significantly even when initialized from the ground truths. The result is illustrated in Fig.~\ref{fig:teaser}.
Our question now becomes: \textbf{What is the reason that prevents the joint optimization of the camera poses with GeNeRFs?} Although a thorough theoretical analysis of the question is difficult due to the high complexity of GeNeRFs, we postulate the potential reasons by 
observing the gradient flow during back-propagation. Particularly, the gradient of $\mathcal{L}_{\text{rgb}}$ with respect to the camera poses can be written as:
\begin{align}
    \label{equ:gradient_generfs}
    \frac{\partial \mathcal{L}_{\text{rgb}}}{\partial \mathbf{P}_j} = 
    &\underbrace{\sum_{i \neq j}^{N} \sum_{\mathbf{u}} \sum_{k}^K 
        \frac{\partial h}{\partial g_k} \cdot \frac{\partial g_k}{\partial \mathbf{f}_i^k} \cdot \frac{\partial \mathbf{f}_i^k}{\partial \mathbf{P}_j} 
    }_{\text{image}\ j\ \text{is one of the nearby views of image}\ i} + \nonumber \\
    &\underbrace{\sum_{m}^M \sum_{\mathbf{u}} \sum_k^K 
        \frac{\partial h}{\partial g_k} \cdot \frac{\partial g_k}{\partial \mathbf{f}_m^k} \cdot \frac{\partial \mathbf{f}_m^k}{\partial \mathbf{P}_j}
    }_{\text{image}\ j\ \text{is the target image}}.
\end{align}
Two problems can arise in the computation of the gradients of $\mathcal{L}_{\text{rgb}}$ given in Eq.~\ref{equ:gradient_generfs}. 1)  \textbf{An image feature can be an outlier}. For example, the sampled pixel of the target view is far away from or missing its correspondences in the nearby views due to occlusion, as illustrated in Fig.~\ref{fig:difficulty_on_optimizing_camera_poses}\subref{fig:generf_outlier_feature}. Without a special design of the network architecture, the aggregation function $f_a(\cdot)$ is not aware of occlusions. Consequently, this causes the two terms $\frac{\partial \mathbf{f}_i^k}{\partial \mathbf{P}_j}$ and $\frac{\partial \mathbf{f}_m^k}{\partial \mathbf{P}_j}$ to be erroneous, and thus causing the final gradient $\frac{\partial \mathcal{L}_{\text{rgb}}}{\partial \mathbf{P}_j}$ to be wrong. 2) \textbf{Non-smooth cost map caused by ResNet-like features.} Fig.~\ref{fig:reset_feature_vs_fpn_feature} (middle) shows an example of the non-smooth cost map from ResNet. Unfortunately, the coarse-to-fine training strategy in BARF to first suppress the high-frequency components and then add them back when the low-frequency components become stabilized is not helpful since most GeNeRFs work directly on the features and do not use positional encodings.

\subsection{DBARF}

Based on the analysis in Sec.~\ref{subsec:on_the_difficulty_of_ba_generfs}, we propose DBARF to jointly optimize the camera poses with GeNeRFs in the following sections. Fig.~\ref{fig:network_architecture} shows our network architecture. To demonstrate our method in detail, we take IBRNet as the GeNeRF method, and we note that it generally does not affect the applicability of our method to other GeNeRFs.

\subsubsection{Camera Poses Optimization}
\label{subsubsec:camera_pose_optimization}

Given a point $\mathbf{X}_i^k$ in the camera frame of target view $i$, IBRNet aggregates features by projecting the point into nearby views:
\begin{equation}
    \Pi ( \mathbf{P}_j, \omega(\mathbf{X}_i^k, \mathbf{P}_i) ) = 
    \mathbf{K}_j \mathbf{P}_j \mathbf{P}_i^{-1} \mathbf{X}_i = \mathbf{K}_j \mathbf{P}_{ij} \mathbf{X}_i^k,
\end{equation}
where $\mathbf{K}_j$ is the intrinsics matrix of image $j$, $\mathbf{P}_{ij} = \mathbf{P}_j \mathbf{P}_i^{-1}$ is the relative 
camera pose from image $i$ to image $j$.

Suppose we have initial camera poses $\mathbf{P}^{\text{init}}$,  we need to first correct the camera poses %at first
before aggregating useful image features. Since the appearances of extracted image features are inconsistent due to inaccurate initial camera poses, an intuitive solution is to construct a cost function that enforces the feature-metric consistency across the target view and all nearby views, \ie:
\begin{equation} \label{equ:point_cost_map}
\small
% \begin{split} 
    \mathcal{C} = 
    \sum_{\mathbf{u}_i} \sum_{j \in \mathcal{N}(i)} \rho \big ( \|
        \chi \big( \mathbf{K}_j \mathbf{P}_{ij} \mathbf{X}_i^k, \mathbf{F}_j \big) - 
        \chi (\mathbf{u}_i, \mathbf{F}_i) \big)
    \|,
% \end{split}
\end{equation}
which has been shown to be more robust than the photometric cost in Eq.~\eqref{equ:photometric_loss} and the keypoint-based bundle 
adjustment~\cite{DBLP:conf/iccv/LindenbergerSLP21}. $\rho(\cdot)$ can be any robust loss function.

However, simply adopting Eq.~\eqref{equ:point_cost_map} to optimize the camera poses
without knowing the outlier distribution to apply a suitable robust loss $\rho(\cdot)$ can give bad results. Furthermore, first-order optimizers can also easily get stuck at bad local minima in our task. Therefore, we seek an approach that can minimize Eq.~\eqref{equ:point_cost_map} while bypassing direct gradient descent. Instead of explicitly taking Eq.~\eqref{equ:point_cost_map} 
as an objective and optimizing camera poses by gradient descent, we implicitly minimize it by taking the feature error as an input to another neural network. Since NeRF randomly samples points in the target view during training, we lose the spatial information of the features when the neural network directly takes Eq.~\eqref{equ:point_cost_map} as input. To alleviate the problem, we sample a patch $\mathcal{S}(\mathbf{u}_i)$ centered on $\mathbf{u}_i$ from the target view for the cost map generation and take the average of the aggregated feature cost map (See Fig.~\ref{fig:deep_optimizer_patch}), \ie:
\begin{equation}
    \label{equ:image_cost_map}
    \small
    \mathcal{C} = 
    \frac{1}{|\mathcal{N}(i)|} \sum_{j \in \mathcal{N}(i)} \|
        \chi \big( \mathbf{K}_j \mathbf{P}_{ij} \mathbf{X}_{\mathcal{S}(\mathbf{u}_i)}, \mathbf{F}_j \big) - 
        \chi (\mathcal{S}(\mathbf{u}_i), \mathbf{F}_i)
    \|,
\end{equation}
where $\mathbf{X}_{\mathcal{S}(\mathbf{u}_i)}$ denotes the patch of 3D points which is computed from a predicted depth map $\mathbf{D}_i$ for the target image $i$ instead of the sampled depth value $Z_{k,i}$ because it is inaccurate. We also do not compute the depth value using Eq.~\eqref{subequ:depth_accumulation} since NeRF does not learn the scene geometry well.

To make the implicit objective smoother to ease the joint training, inspired by BANet~\cite{DBLP:conf/iclr/TangT19}, 
we adopt the FPN (Feature Pyramid Network)~\cite{DBLP:conf/cvpr/LinDGHHB17} 
as our feature backbone. Given a cost feature map in Eq.~\eqref{equ:image_cost_map}, we aim at updating the relative camera poses $\mathbf{P}_{ij}$ and the depth map $\mathbf{D}_i$.

Following the RAFT-like~\cite{DBLP:conf/eccv/TeedD20,DBLP:journals/corr/abs-2103-13201,DBLP:conf/ijcai/Teed021}
architecture, we adopt a recurrent GRU to predict the camera poses and depth map. Given initial camera poses $\mathbf{P}_{ij}^0$ and depth $\mathbf{D}_i^0$, we compute an initial cost map $\mathcal{C}^0$ using Eq.~\eqref{equ:image_cost_map}. We then use a GRU to predict the relative camera pose correction $\Delta \mathbf{P}_{ij}$ and depth correction $\Delta \mathbf{D}_k$ at the current iteration $t$, and update the camera poses and depth, respectively, as:
\begin{align}
    \label{equ:pose_depth_update}
    %& 
    \mathbf{P}_{ij}^{t+1} = \mathbf{P}_{ij}^t + \Delta \mathbf{P}_{ij}, \quad %\nonumber %\\
    %& 
    \mathbf{D}_i^{t+1} = \mathbf{D}_i^t + \Delta \mathbf{D}_i.
\end{align}
During training, $\mathbf{P}_{ij}^0$ and $\mathbf{D}_i^0$  are randomly initialized and  Eq.~\eqref{equ:pose_depth_update} is executed for a fixed $t$ iteration. 
Note that after each iteration, the cost map $\mathcal{C}^{t+1}$ is updated by taking the current relative poses and depth map as input. Stage A in Fig.~\ref{fig:network_architecture} illustrates the recurrent updating step.

\subsubsection{Scene Graph: Nearby Views Selection}
Existing GeNeRFs aggregate features from nearby views by selecting the nearest top-$k$ nearby views with the known absolute camera poses. 
Since the absolute camera poses are not given in our setting, we select the nearby views using a scene graph. A scene graph records neighbors of a target view $\mathbf{I}_i$. To construct the scene graph, we extract keypoints for each image using SuperPoint~\cite{DBLP:conf/cvpr/DeToneMR18} and obtain feature matches for each candidate image pair using SuperGlue~\cite{DBLP:conf/cvpr/SarlinDMR20}. 
Wrong feature matches are filtered by checking the epipolar constraints~\cite{DBLP:books/daglib/0015576}. Two images become neighbors when they share enough image keypoint matches. We simply select nearby views by sorting their neighbors according to the number of inlier matches in descending order. The scene graph construction only needs to be executed once for each scene and thus is a preprocessing step.

\subsection{Training Objectives}
\label{subsec:training}

For depth optimization, we adopt the edge-aware depth map smoothness loss in \cite{DBLP:conf/iccv/GodardAFB19} for self-supervised 
depth prediction, which can penalize changes where the original image is smooth:
\begin{equation}
    \label{equ:smoothness_loss}
    \mathcal{L}_{\text{depth}} = |\partial_x \mathbf{D}| \exp^{-|\partial_x \mathbf{I}|} + |\partial_y \mathbf{D}| \exp^{-|\partial_y \mathbf{I}|},
\end{equation}
where $\partial_x$ and $\partial_y$ are the image gradients.

For camera poses optimization, we adopt the warped photometric loss~\cite{DBLP:journals/corr/abs-2103-13201} for self-supervised 
pose optimization:
\begin{equation}
    \label{equ:warped_photometric_loss}
    \mathcal{L}_{\text{photo}} = \frac{1}{| \mathcal{N}_i |} \sum_{j \in \mathcal{N}_i} \big(
                                \alpha \frac{1-\text{ssim} (\mathbf{I}_i^{'} - \mathbf{I}_i)}{2} + 
                                (1-\alpha) \| \mathbf{I}_i^{'} - \mathbf{I}_i \| \big),
\end{equation}
where $\mathbf{I}_i^{'}$ is warped from nearby image $j$ to the target image $i$, $\text{ssim}$ is the structural similarity 
loss~\cite{DBLP:journals/tip/WangBSS04}.

For GeNeRF, we use the same loss of Eq.~\eqref{equ:photometric_loss}. Finally, our final loss function is defined as:
\begin{equation}
    \label{equ:final_loss}
    \mathcal{L}_{\text{final}} = 2^{\beta \cdot t} (\mathcal{L}_{\text{depth}} + \mathcal{L}_{\text{photo}}) + 
        (1-2^{\beta \cdot t}) \mathcal{L}_{\text{rgb}},
\end{equation}
where $\beta=-1e5$, $t$ is the current training iteration number.

%%%%%%%%%%%%%%%%%%%%%%%%%%%%%%%%%%%%%%%%%%%%%%%%%%%%%%%%%%%%% Experiments %%%%%%%%%%%%%%%%%%%%%%%%%%%%%%%%%%%%%%%%%%%%%%%%%%%%%%%%%%
\section{Experiments}
\label{sec:experiments}

\paragraph{Training Datasets.}
We pretrain IBRNet and our method on the 63 scenes of the self-collected datasets from IBRNet~\cite{DBLP:conf/cvpr/WangWGSZBMSF21}, the 33 real scenes captured by a handheld headphone from LLFF~\cite{DBLP:journals/tog/MildenhallSCKRN19} with the ground truth camera poses obtained from COLMAP, and 20 indoor scenes from the ScanNet dataset~\cite{DBLP:conf/cvpr/DaiCSHFN17} with ground truth camera poses provided by BundleFusion~\cite{DBLP:journals/tog/DaiNZIT17}. The ground truth camera poses are provided by IBRNet, but not used in our method.

\vspace{-3mm}
\paragraph{Evaluation Datasets.}
We evaluate BARF, IBRNet, and our method on the LLFF dataset~\cite{DBLP:journals/tog/MildenhallSCKRN19} and the 
ScanNet dataset~\cite{DBLP:conf/cvpr/DaiCSHFN17}. For IBRNet and our method, the evaluated scenes are not used during pre-training. For BARF and GARF, we train and evaluate them on the same scene in $200,000$ iterations. 1/8th and 1/20th of the images are respectively held out for testing on LLFF and ScanNet while others are reserved for finetuning. More scenes are evaluated in our supplementary materials.

\vspace{-3mm}
\paragraph{Implementation Details.} We adopt IBRNet~\cite{DBLP:conf/cvpr/WangWGSZBMSF21} as the GeNeRF implementation and DRO~\cite{DBLP:journals/corr/abs-2103-13201} as the pose optimizer. Our method and IBRNet are both trained on a single 24G NVIDIA RTX A5000 GPU. We train our method end-to-end using Adam~\cite{DBLP:journals/corr/KingmaB14} with a learning rate of $1e-3$ for the feature extractor, $5e-4$ for GeNeRF, and $2e-4$ for pose optimizer during pretraining. For fine-tuning, the learning rate is $5e-4$ for the feature extractor, $2e-4$ for GeNeRF, and $1e-5$ for the pose optimizer. We pretrain IBRNet in $250,000$ iterations and our method in $200,000$ iterations and finetune both IBRNet and our method in $60,000$ iterations. During pretraining, for our method, we only select 5 nearby views for pose correction and novel view rendering for efficiency. During fine-tuning and evaluation, we select 10 nearby views for both our method and IBRNet. The camera poses are updated by 4 iterations in a batch. Note that vanilla NeRF~\cite{DBLP:conf/eccv/MildenhallSTBRN20} and IBRNet~\cite{DBLP:conf/cvpr/WangWGSZBMSF21} use a coarse network and a fine network to predict density value and color. However, BARF~\cite{DBLP:conf/iccv/LinM0L21} and GARF~\cite{DBLP:journals/corr/abs-2204-05735} use a single coarse network. To make a fair comparison to them, we only train a coarse network for IBRNet and our method.

\begin{figure*}[htbp]
    \centering
    % \subfloat {
    %     \includegraphics[width=0.95\linewidth]{llff_flower_render}
    %     \label{fig:llff_flower_render}
    % } \\
    \subfloat {
        \includegraphics[width=0.95\linewidth]{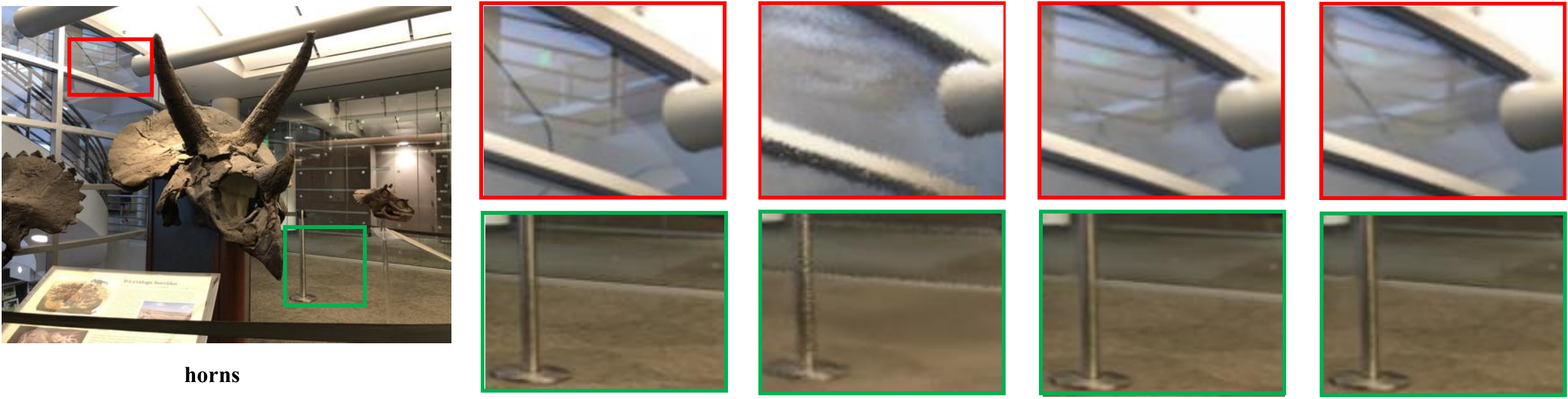}
        \label{fig:llff_horns_render}
    } \\
    \subfloat {
        \includegraphics[width=0.95\linewidth]{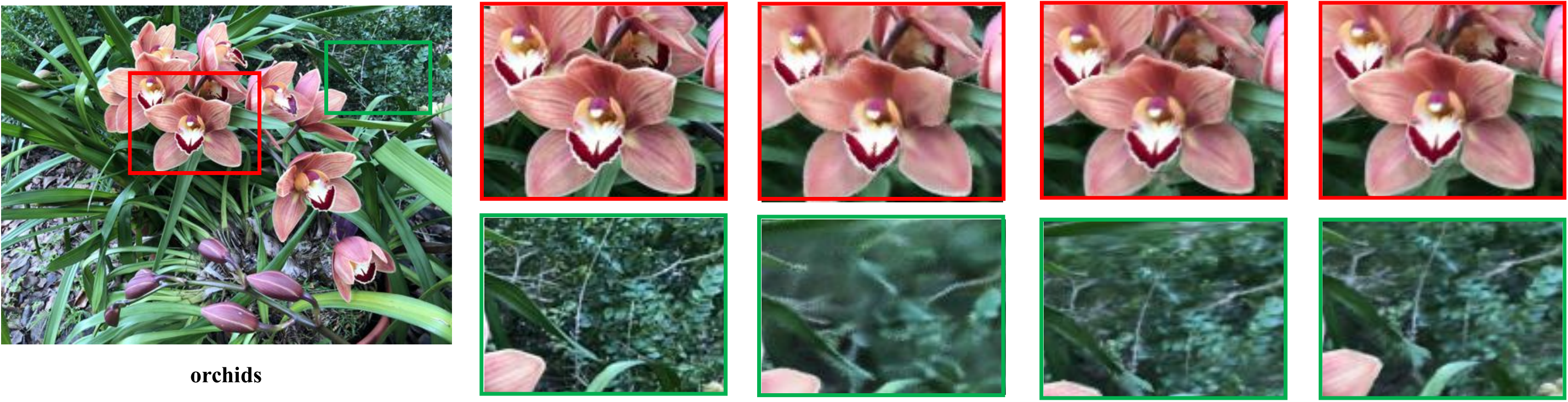}
        \label{fig:llff_orchids_render}
    } \\
    % \subfloat {
    %     \includegraphics[width=0.95\linewidth]{llff_leaves_render}
    %     \label{fig:llff_leaves_render}
    % } \\
    \subfloat {
        \includegraphics[width=0.95\linewidth]{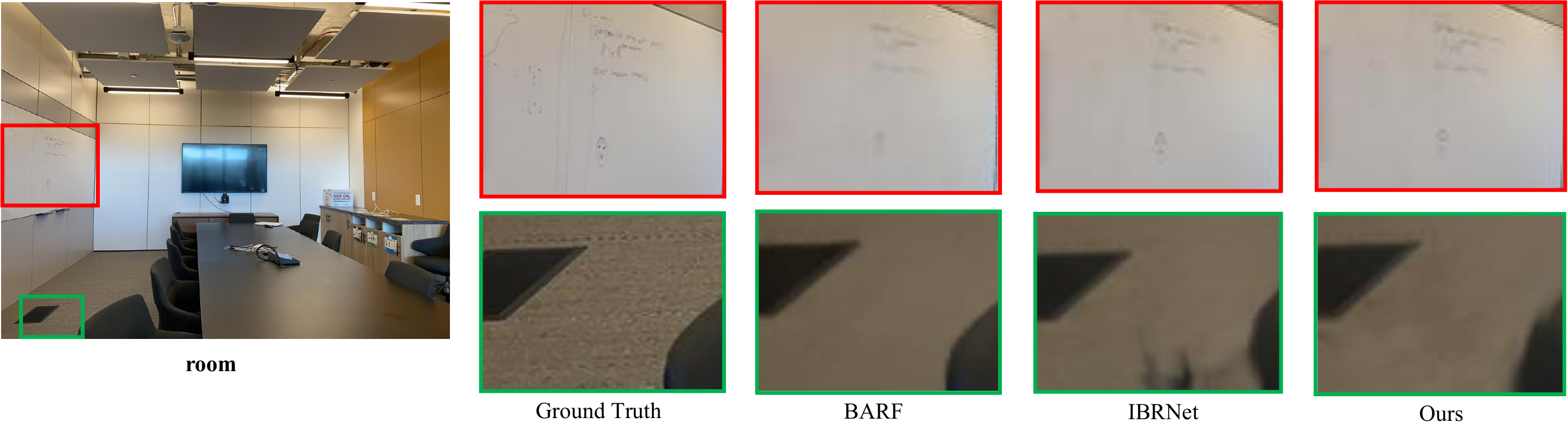}
        \label{fig:llff_room_render}
    } \\
    \vspace{-3mm}
    \caption{\textbf{The qualitative results on LLFF forward-facing dataset~\cite{DBLP:journals/tog/MildenhallSCKRN19}}.
    We show the finetuned results for IBRNet and Ours.}
    \vspace{-2mm}
    \label{fig:llff_render}
\end{figure*}

\begin{figure*}[htbp]
    \centering
    \includegraphics[width=0.98\linewidth]{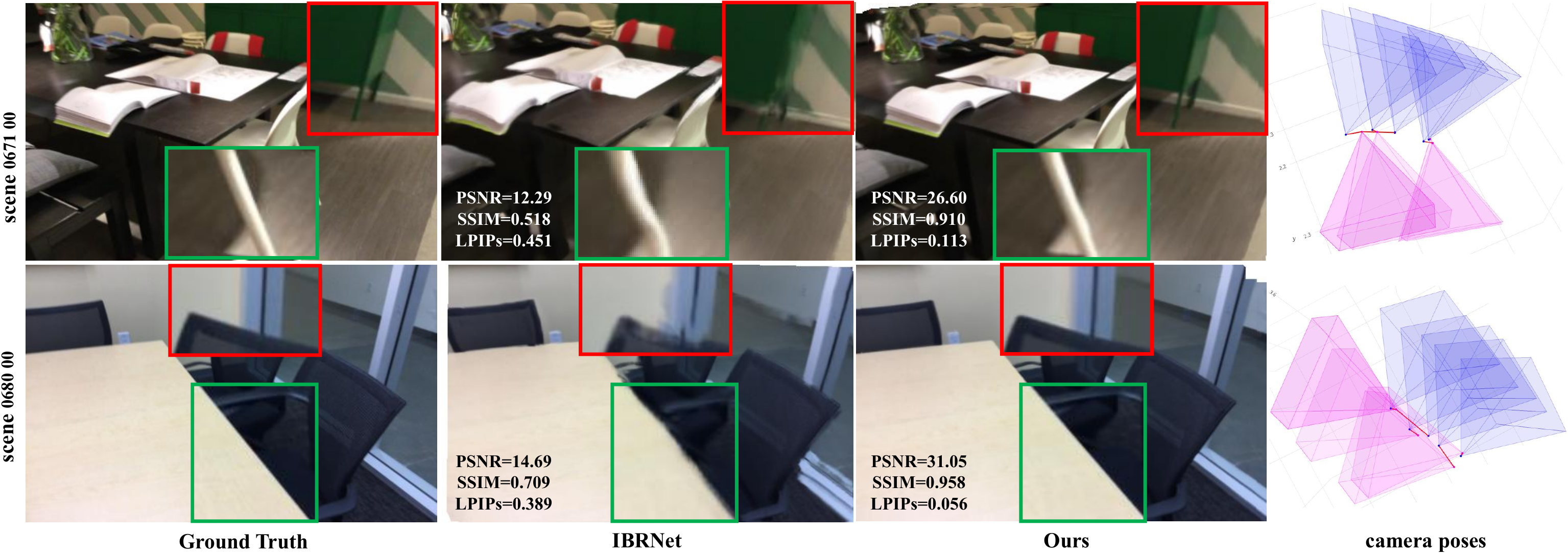}
    \vspace{-3mm}
    \caption{\textbf{The qualitative results on ScanNet dataset~\cite{DBLP:conf/cvpr/DaiCSHFN17}}.
    We show the finetuned results for IBRNet and Ours. Red and blue are  
    the pseudo ground truth (used by IBRNet) and the predicted camera poses of our method, respectively. }
    \label{fig:scannet_render_pose}
    \vspace{-5mm}
\end{figure*}

\subsection{Experimental Results}
We evaluated both the rendering quality for novel view synthesis and pose accuracy of our method.
The code of GARF~\cite{DBLP:journals/corr/abs-2204-05735} is not publicly available 
during this work, and thus we cite the quantitative results from the original paper.

\begin{figure*}[htbp]
    \centering
    \includegraphics[width=0.98\linewidth]{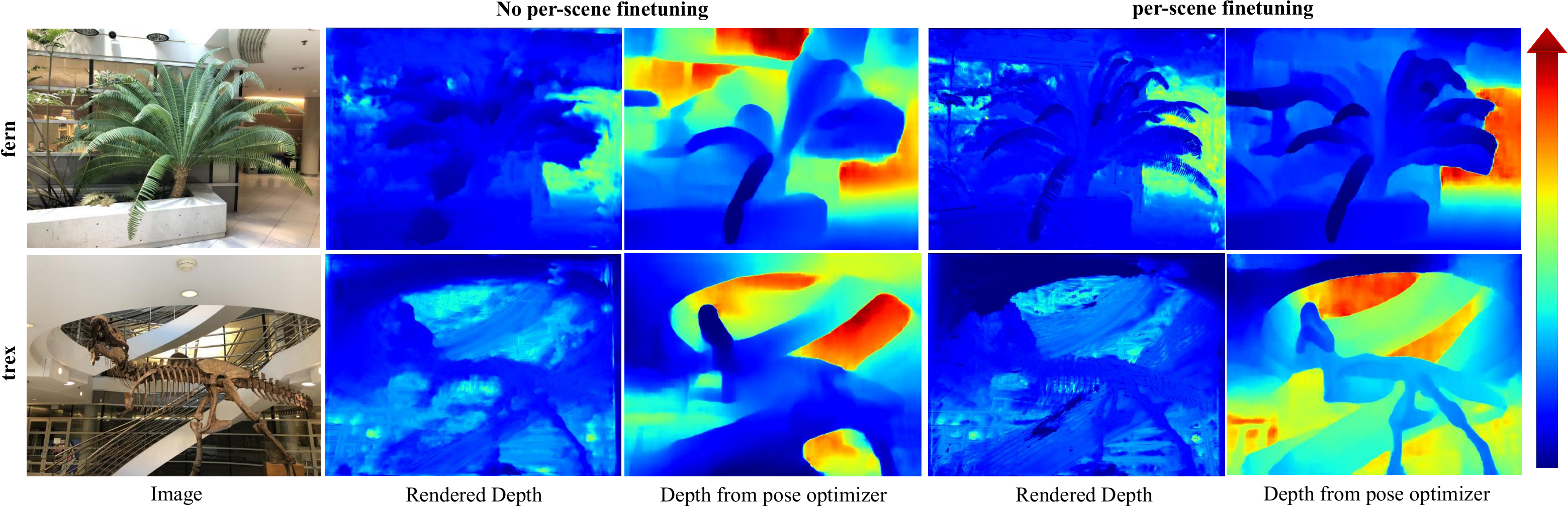}
    \vspace{-2mm}
    \caption{\textbf{Depth maps on LLFF forward-facing dataset~\cite{DBLP:journals/tog/MildenhallSCKRN19}}. The rendered depth is computed from our GeNeRF %part 
    after fine-tuning.}
    \label{fig:llff_depth}
    \vspace{-1mm}
\end{figure*}

% ------------------------------------------ Table ---------------------------------------------------
\begin{table*}[htbp]
    \centering
    \resizebox{1.0\textwidth}{!}{
      \begin{tabular}{c | c c c c c c | c c c c c c | c c c c c c }
        \toprule
  
        \multirow{3}{*}{Scenes}  &
        \multicolumn{6}{c|}{\textbf{PSNR}\ $\uparrow$} &
        \multicolumn{6}{c|}{\textbf{SSIM}\ $\uparrow$} &  
        \multicolumn{6}{c}{\textbf{LPIPS}\ $\downarrow$} \\
        
        \cmidrule(r){2-7} \cmidrule(r){8-13} \cmidrule(r){14-19}
        & \multirow{2}{*}{BARF~\cite{DBLP:conf/iccv/LinM0L21}} & \multirow{2}{*}{GARF~\cite{DBLP:journals/corr/abs-2204-05735}} 
        & \multicolumn{2}{c}{IBRNet~\cite{DBLP:conf/cvpr/WangWGSZBMSF21}} & \multicolumn{2}{c|}{Ours}
        
        & \multirow{2}{*}{BARF~\cite{DBLP:conf/iccv/LinM0L21}} & \multirow{2}{*}{GARF~\cite{DBLP:journals/corr/abs-2204-05735}} 
        & \multicolumn{2}{c}{IBRNet~\cite{DBLP:conf/cvpr/WangWGSZBMSF21}} & \multicolumn{2}{c|}{Ours}
        
        & \multirow{2}{*}{BARF~\cite{DBLP:conf/iccv/LinM0L21}} & \multirow{2}{*}{GARF~\cite{DBLP:journals/corr/abs-2204-05735}} 
        & \multicolumn{2}{c}{IBRNet~\cite{DBLP:conf/cvpr/WangWGSZBMSF21}} & \multicolumn{2}{c}{Ours} \\
        
        \cmidrule(r){4-5} \cmidrule(r){6-7} \cmidrule(r){10-11} \cmidrule(r){12-13} \cmidrule(r){16-17} \cmidrule(r){18-19} & 
        & & \text{\XSolidBrush} & \text{\CheckmarkBold} & \text{\XSolidBrush} & \text{\CheckmarkBold} & 
        & & \text{\XSolidBrush} & \text{\CheckmarkBold} & \text{\XSolidBrush} & \text{\CheckmarkBold} &
        & & \text{\XSolidBrush} & \text{\CheckmarkBold} & \text{\XSolidBrush} & \text{\CheckmarkBold} \\

    \midrule
  
        fern & 23.79 & 24.51 & 23.61 & 25.56 & 23.12 & \textbf{25.97}
             & 0.710 & 0.740 & 0.743 & 0.825 & 0.724 & \textbf{0.840} 
             & 0.311 & 0.290 & 0.240 & 0.139 & 0.277 & \textbf{0.120}  \\

        flower & 23.37 & \textbf{26.40} & 22.92 & 23.94 & 21.89 & 23.95 
               & 0.698 & 0.790 & 0.849 & 0.895 & 0.793 & \textbf{0.895}
               & 0.211 & 0.110 & 0.123 & 0.074 & 0.176 & \textbf{0.074}  \\
        
        fortress & 29.08 & 29.09 & 29.05 & 31.18 & 28.13 & \textbf{31.43}
                 & 0.823 & 0.820 & 0.850 & 0.918 & 0.820 & \textbf{0.918}
                 & 0.132 & 0.150 & 0.087 & 0.046 & 0.126 & \textbf{0.046} \\
        
        horns & 22.78 & 23.03 & 24.96 & \textbf{28.46} & 24.17 & 27.51
              & 0.727 & 0.730 & 0.831 & \textbf{0.913} & 0.799 & 0.903
              & 0.298 & 0.290 & 0.144 & \textbf{0.070} & 0.194 & 0.076  \\

        leaves & 18.78 & 19.72 & 19.03 & \textbf{21.28} & 18.85 & 20.32
               & 0.537 & 0.610 & 0.737 & \textbf{0.807} & 0.649 & 0.758
               & 0.353 & 0.270 & 0.289 & \textbf{0.137} & 0.313 & 0.156  \\

        orchids & 19.45 & 19.37 & 18.52 & \textbf{20.83} & 17.78 & 20.26
                & 0.574 & 0.570 & 0.573 & \textbf{0.722} & 0.506 & 0.693
                & 0.291 & 0.260 & 0.259 & \textbf{0.142} & 0.352 & 0.151  \\
        
        room & \textbf{31.95} & 31.90 & 28.81 & 31.05 & 27.50 & 31.09
             & 0.940 & 0.940 & 0.926 & \textbf{0.950} & 0.901 & 0.947
             & 0.099 & 0.130 & 0.099 & \textbf{0.060} & 0.142 & 0.063  \\
        
        trex & 22.55 & 22.86 & 23.51 & \textbf{26.52} & 22.70 & 22.82
             & 0.767 & 0.800 & 0.818 & \textbf{0.905} & 0.783 & 0.848
             & 0.206 & 0.190 & 0.160 & \textbf{0.074} & 0.207 & 0.120  \\
        \bottomrule
      \end{tabular}
    }
    \vspace{-2mm}
    \caption{Quantitative results of novel view synthesis on LLFF~\cite{DBLP:journals/tog/MildenhallSCKRN19} forward-facing dataset. For 
             IBRNet~\cite{DBLP:conf/cvpr/WangWGSZBMSF21} and our method, the results with (\CheckmarkBold) and without (\XSolidBrush) 
             per-scene fine-tuning are given.}
    \vspace{-4mm}
    \label{table:llff_quantitative_nvs}
\end{table*}

\vspace{-3mm}
\paragraph{Novel View Synthesis.}
We use PSNR, SSIM~\cite{DBLP:journals/tip/WangBSS04} and LPIPS~\cite{DBLP:conf/cvpr/ZhangIESW18} as the metrics for novel view synthesis.
The quantitative results are shown in Table~\ref{table:llff_quantitative_nvs}. As we can see, the rendering quality of our method surpasses both 
BARF and GARF, and we even outperform IBRNet on the fern, flower, and fortress scenes with the unfair advantage that IBRNet has known camera poses (ours does not). The qualitative results on the LLFF dataset are given in Fig.~\ref{fig:llff_render}. For IBRNet and our method, we show the per-scene finetuned visual results. 
For the scenes of \textit{horns} and \textit{orchids}, our method even renders images with higher quality than IBRNet. For the \textit{room} scene, we can observe an obvious artifact for IBRNet (floor in the green zoomed-in area). This validated the effectiveness of our method.
We also present the rendering results of IBRNet and our method on the ScanNet dataset in Fig.~\ref{fig:scannet_render_pose}. 
Our method renders much better results than IBRNet. Furthermore, the differences in the camera poses visualized in Fig.~\ref{fig:scannet_render_pose} indicate ground truth camera poses are not accurate. Refer to our supplementary for more results.

\vspace{-3mm}
\paragraph{Pose Accuracy.}
Since our DBARF does not recover absolute camera poses, we measure the accuracy of the predicted relative camera poses. 
Specifically, for each test scene, we select one batch of nearby views for all images and then recover the relative poses 
from the target view to each nearby view. The target view's camera pose is set to identity, then we estimate a similarity 
transformation to align all camera poses in that batch to ground truth by Umeyama~\cite{DBLP:journals/pami/Umeyama91}. 
The pose accuracy is measured by taking the average of all pose errors between the predicted relative poses and ground truth 
camera poses. The quantitative results are given in Table.~\ref{table:llff_quantitative_pose}.

% ------------------------------------------ Table ---------------------------------------------------
\begin{table}[htbp]
    \centering
    \resizebox{0.5\textwidth}{!}{
      \begin{tabular}{c c c c c c c c c }
        \toprule
  
        \multirow{1}{*}{Scenes}  &
        \multicolumn{1}{c}{\textbf{fern}} &
        \multicolumn{1}{c}{\textbf{flower}} &  
        \multicolumn{1}{c}{\textbf{fortress}} &
        \multicolumn{1}{c}{\textbf{horns}} &
        \multicolumn{1}{c}{\textbf{leaves}} &
        \multicolumn{1}{c}{\textbf{orchids}} & 
        \multicolumn{1}{c}{\textbf{room}} &
        \multicolumn{1}{c}{\textbf{trex}} \\

        \midrule
  
        Rotation (\XSolidBrush) & 9.96 & 16.74 & 2.18  & 6.08 & 12.98 & 5.90 & 8.76 & 10.09 \\
        Rotation (\CheckmarkBold) & 0.89  & 1.39  & 0.59 & 0.82 & 4.63  & 1.164 & 0.53 & 1.06 \\
        
        \hline
        
        translation (\XSolidBrush) & 2.00 & 1.56  & 1.06 & 2.45 & 2.56 & 5.13 & 5.48 & 8.05 \\
        translation (\CheckmarkBold) & 0.34 & 0.32 & 0.23 & 0.29 & 0.85 & 0.57 & 0.36 & 0.46 \\

        \bottomrule
      \end{tabular}
    }
    \vspace{-2mm}
    \caption{Quantitative results of camera pose accuracy on LLFF~\cite{DBLP:journals/tog/MildenhallSCKRN19} forward-facing dataset.
             Rotation (degree) and translation (scaled by $10^{2}$, without known absolute scale) errors with (\CheckmarkBold) and without 
             (\XSolidBrush) per-scene fine-tuning are given. }
    \vspace{-5mm}
    \label{table:llff_quantitative_pose}
\end{table}

\vspace{-3mm}
\paragraph{Discussion of the Generalization of DBARF.} To ablate the generalization ability of our method, we show the results of our 
method with and without fine-tuning in Tables.~\ref{table:llff_quantitative_nvs} and~\ref{table:llff_quantitative_pose}. We can 
observe that our method surpasses BARF and GARF on novel view synthesis even without per-scene fine-tuning. For pose accuracy, the rotation error of our method is less than $13^{\circ}$ for most of the scenes in the LLFF dataset without fine-tuning, which 
is much cheaper than per-scene training from scratch. Our rotation error is less than $1.5^{\circ}$ (except for the leaves scene) with fine-tuning. 
This proves that our method is generalizable across scenes. We also argue that the camera poses computed by COLMAP are only pseudo ground truth. Our camera poses are better than COLMAP since the rendering quality of our DBARF is better than IBRNet on the \textit{fern}, \textit{flower}, and \textit{fortress} scenes.

\vspace{-3mm}
\paragraph{Qualitative Analysis of Depth Maps.} In Fig.~\ref{fig:llff_depth}, we present the depth maps computed from NeRF in Eq.~\eqref{subequ:depth_accumulation} (\ie rendered depth maps), and those predicted by our pose optimizer. It can be observed that the depth maps from our pose optimizer are better than those from NeRF, which validates the rationalization of our analysis in Sec.~\ref{subsubsec:camera_pose_optimization}, \ie utilizing the rendered depth map from NeRF to compute the cost map may cause our DBARF to diverge. However, we can observe that while the pose optimizer generates a smoother depth map, NeRF can recover more accurate depth at scene details, especially the thin structures. We believe both depth maps can be improved under self-supervision: NeRF can learn better scene geometry, and the pose optimizer can predict more accurate camera poses with better-quality depth maps.

\vspace{-1mm}
\section{Conclusion}
\label{sec:conclusion}

We analyzed the difficulties of bundle adjusting GeNeRFs, where existing methods such as BARF and its variants cannot work. Based on the analysis, we 
proposed DBARF that can bundle adjust camera poses with GeNeRFs, and can also be jointly trained with GeNeRFs end-to-end without ground truth camera poses. In contrast to BARF and GARF, which require expensive per-scene optimization and good initial camera poses, our proposed DBARF is generalizable across scenes and does require any initialization of the camera poses.

\noindent \textbf{Acknowledgement.} This research/project is supported by the National Research Foundation Singapore and DSO National Laboratories under the AI Singapore Programme (Award Number: AISG2-RP-2020-016), and the Tier 2 grant MOE-T2EP20120-0011 from the Singapore 
Ministry of Education.

\section{APPENDIX}

\subsection{Additional Implementation Details}

\begin{figure}[htbp]
    \centering

    \includegraphics[width=0.99\linewidth]{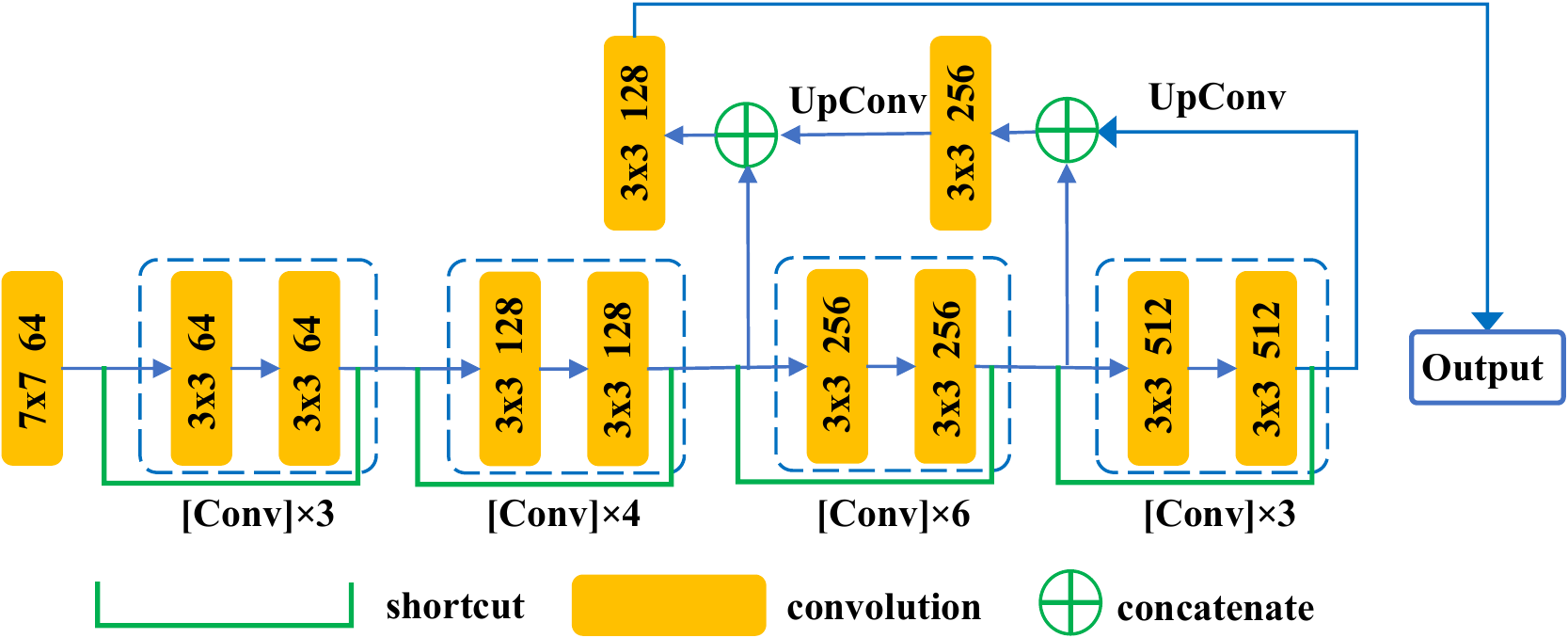}

    \caption{Feature extraction network architecture. The max-pooling layer after the first convolution layer is omitted. Each residual 
             block is a standard block in ResNet~\cite{DBLP:conf/cvpr/HeZRS16}. "Conv" denotes convolution, "UpConv" denotes up-convolution.}
    \label{fig:feature_extraction_backbone}
\end{figure}

\paragraph{Feature extraction network architecture.} We use ResNet34~\cite{DBLP:conf/cvpr/HeZRS16} as our feature extraction network 
backbone. The detailed network architecture is given in Fig.~\ref{fig:feature_extraction_backbone}. Note that after up-convolution, the 
shape of resulted feature map may not be the same as the shape of the feature map to be concatenated. In this case, we use bilinear 
interpolation to rescale them to the same shape.

\paragraph{Depth sampling in GeNeRF.} For both IBRNet and our method, we sample 64 inverse-depth 
samples uniformly. For BARF~\cite{DBLP:conf/iccv/LinM0L21} and GARF~\cite{DBLP:journals/corr/abs-2204-05735}, 
the number of inverse depth samples is 128.

\subsection{Results on LLFF Dataset}

\begin{figure*}[htbp]
    \centering
    % \subfloat {
    %     \includegraphics[width=0.99\linewidth]{llff_flower_render_depth}
    %     \label{fig:llff_flower_render_depth}
    % } \\
    % \subfloat {
    %     \includegraphics[width=0.99\linewidth]{llff_fortress_render_depth}
    %     \label{fig:llff_fortress_render_depth}
    % } \\
    \subfloat {
        \includegraphics[width=0.99\linewidth]{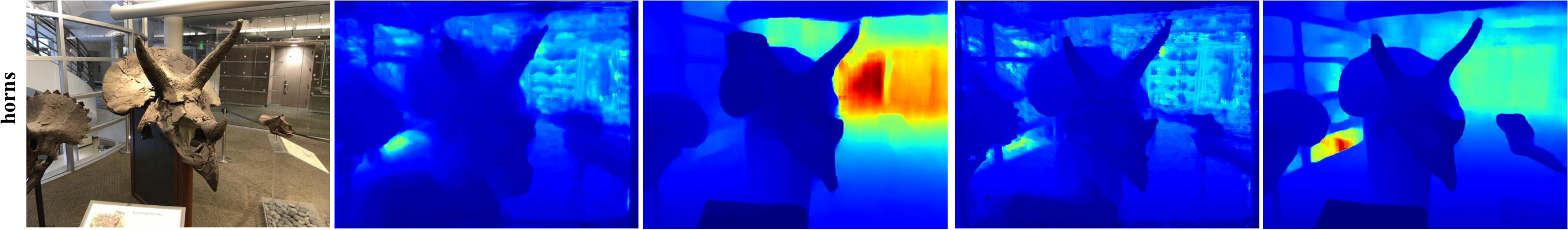}
        \label{fig:llff_horns_render_depth}
    } \\
    \subfloat {
        \includegraphics[width=0.99\linewidth]{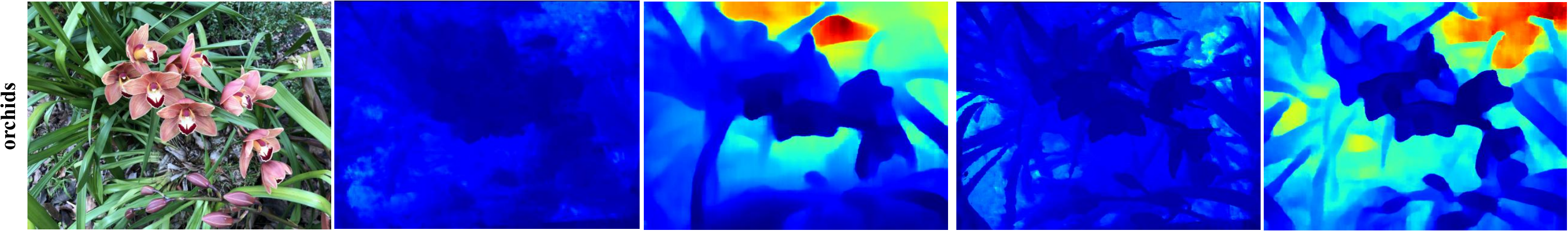}
        \label{fig:llff_orchids_render_depth}
    } \\
    \subfloat {
        \includegraphics[width=0.99\linewidth]{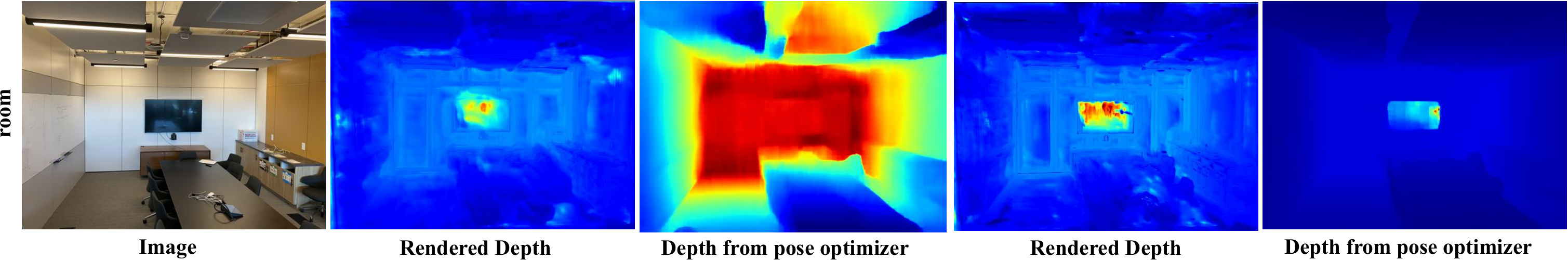}
        \label{fig:llff_room_render_depth}
    }
    \vspace{-3mm}
    \caption{\textbf{Depth visualization on LLFF forward-facing dataset~\cite{DBLP:journals/tog/MildenhallSCKRN19}}.
    We show the finetuned results for IBRNet and Ours.}
    \vspace{-1mm}
    \label{fig:llff_render_depth}
\end{figure*}

We present more qualitative results of the LLFF forward-facing dataset~\cite{DBLP:journals/tog/MildenhallSCKRN19} in 
Fig.~\ref{fig:llff_render} and more geometry visualization results in Fig.~\ref{fig:llff_render_depth}.

\begin{figure*}[htbp]
    \centering
    \subfloat {
        \includegraphics[width=0.99\linewidth]{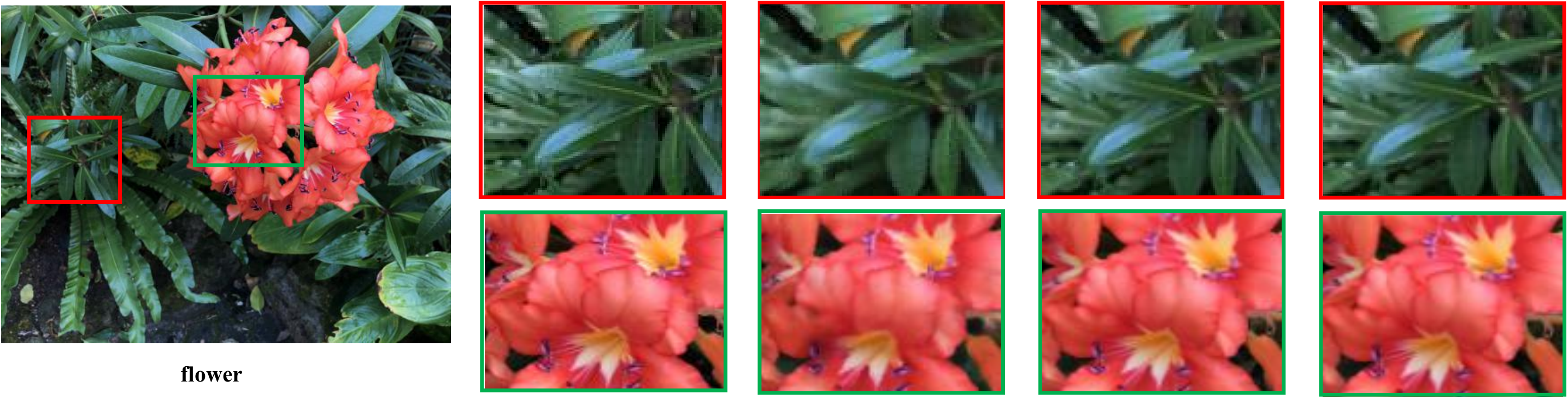}
        \label{fig:llff_flower_render}
    } \\
    \subfloat {
        \includegraphics[width=0.99\linewidth]{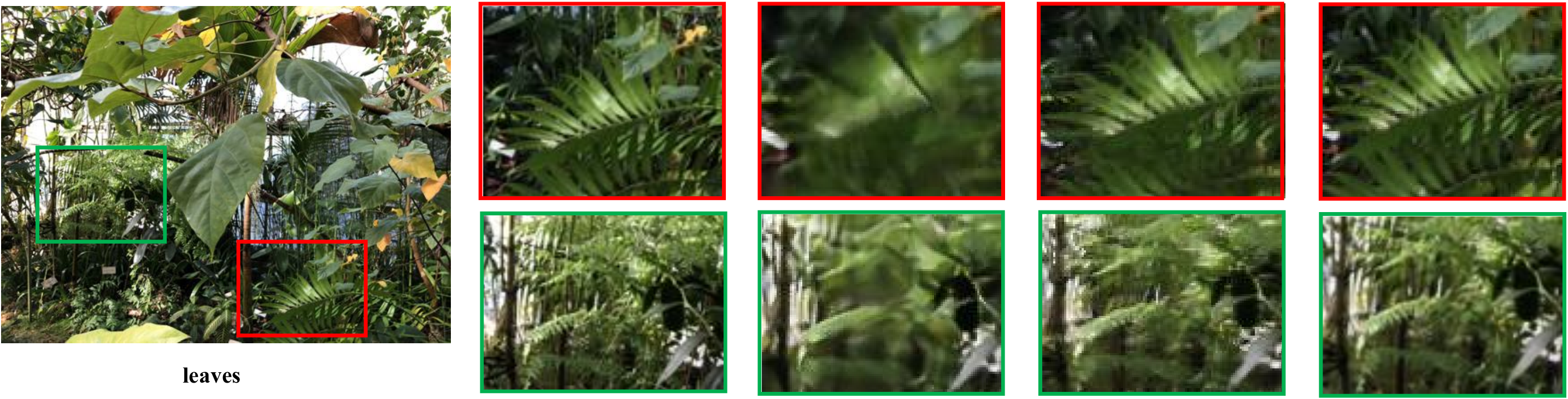}
        \label{fig:llff_leaves_render}
    } \\
    \subfloat {
        \includegraphics[width=0.99\linewidth]{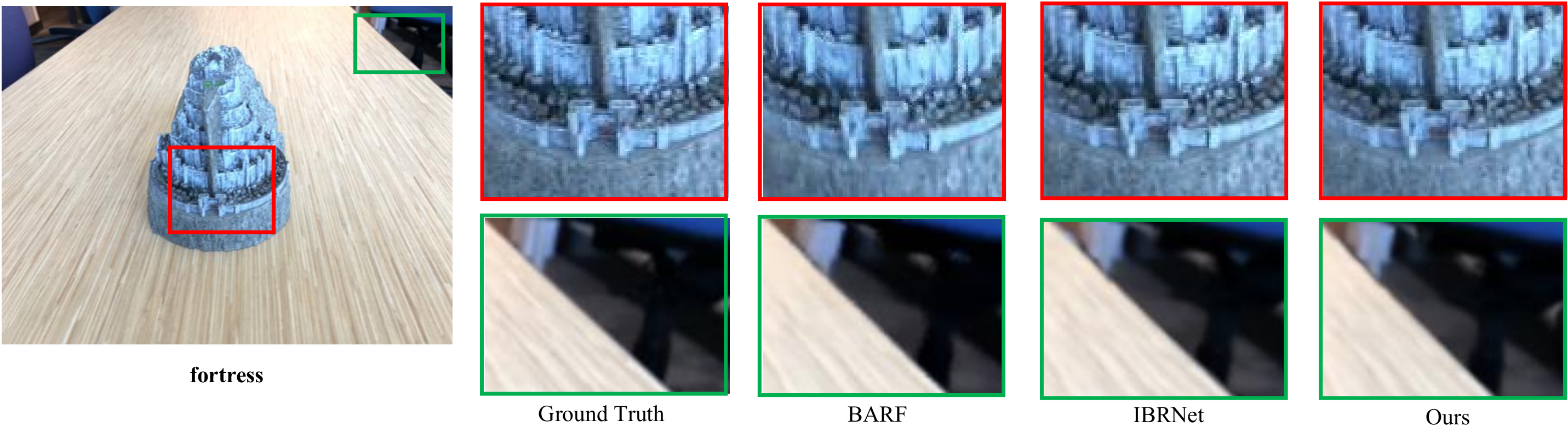}
        \label{fig:llff_fortress_render}
    }
    \vspace{-3mm}
    \caption{\textbf{The qualitative results on LLFF forward-facing dataset~\cite{DBLP:journals/tog/MildenhallSCKRN19}}.
    We show the finetuned results for IBRNet and Ours.}
    \vspace{-3mm}
    \label{fig:llff_render}
\end{figure*}

\subsection{Results on IBRNet's Self-Collected Dataset}

\begin{figure*}[htbp]
    \centering
    \includegraphics[width=0.99\linewidth]{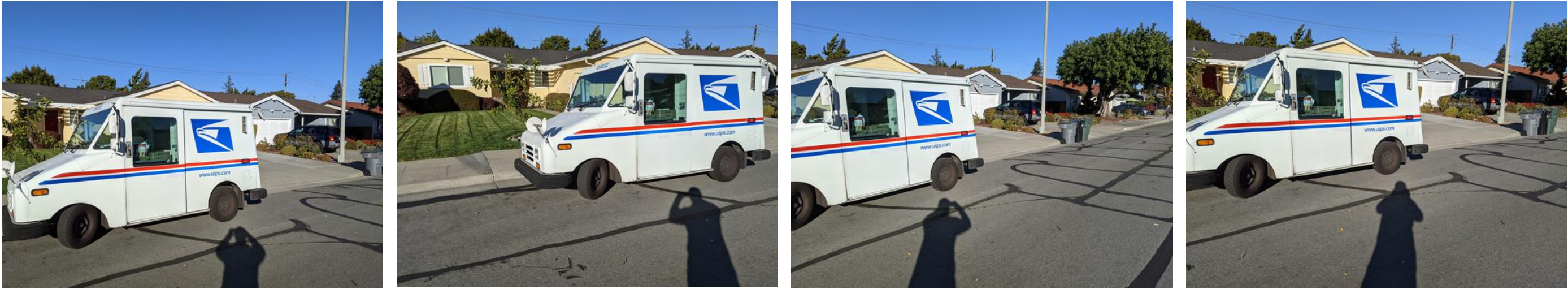}
    
    \caption{A moving human shadow on the `usps van' scene.}
    \label{fig:usps_van}
\end{figure*}

\begin{figure*}[htbp]
    \centering
    \subfloat {
        \includegraphics[width=0.95\linewidth]{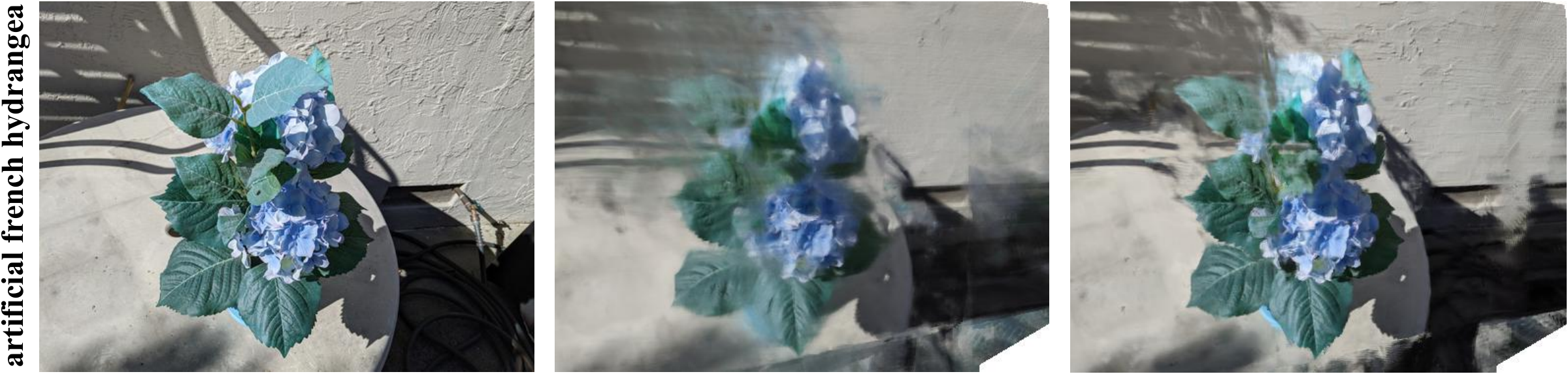}
        \label{fig:ibrnet_collected_artificial_french_hydrangea}
    } \\ 
    \subfloat {
        \includegraphics[width=0.95\linewidth]{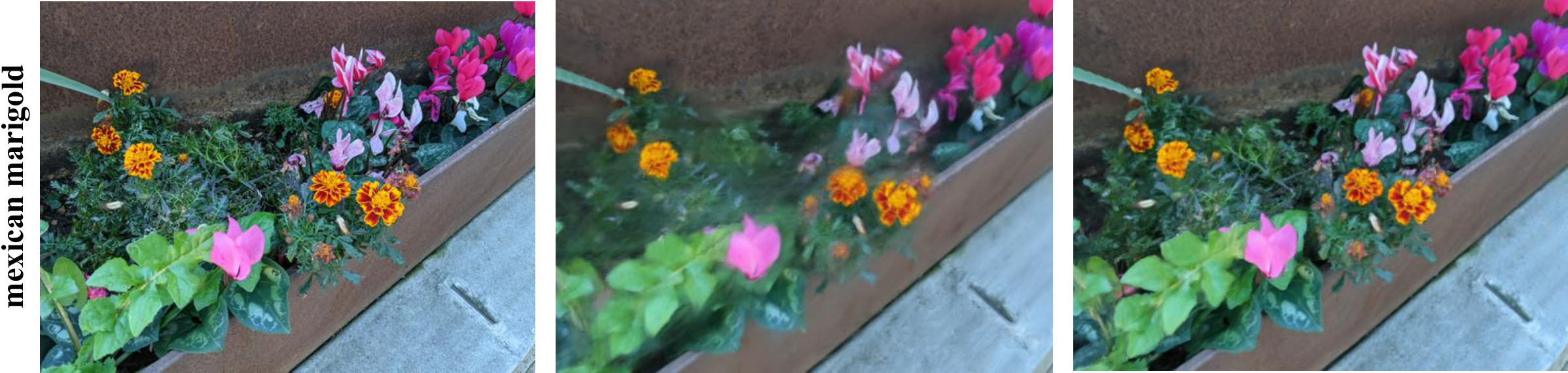}
        \label{fig:ibrnet_collected_mexican_marigold}
    } \\
    \subfloat {
        \includegraphics[width=0.95\linewidth]{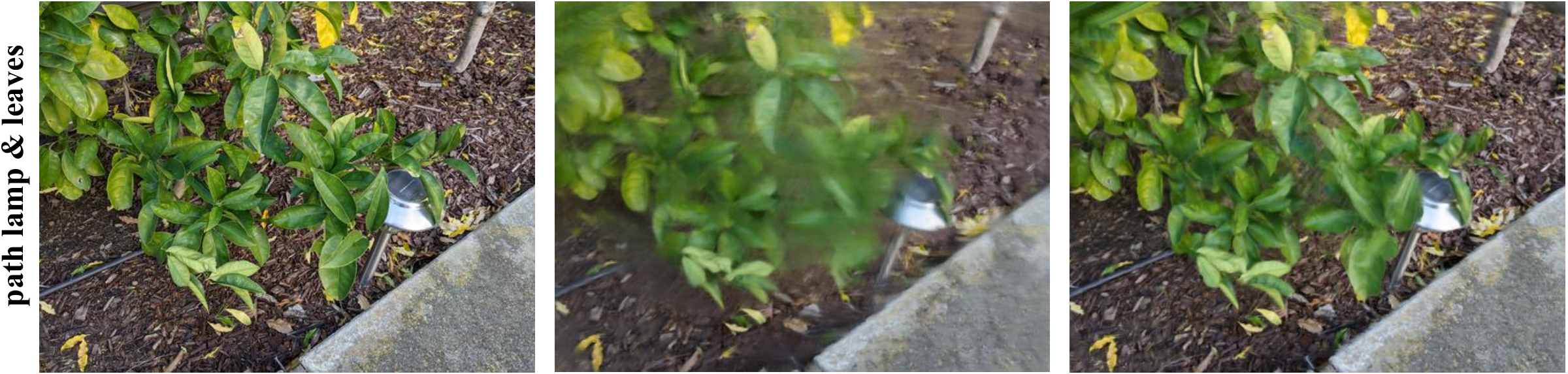}
        \label{fig:ibrnet_collected_path_lamp_leaves}
    } \\
    \subfloat {
        \includegraphics[width=0.95\linewidth]{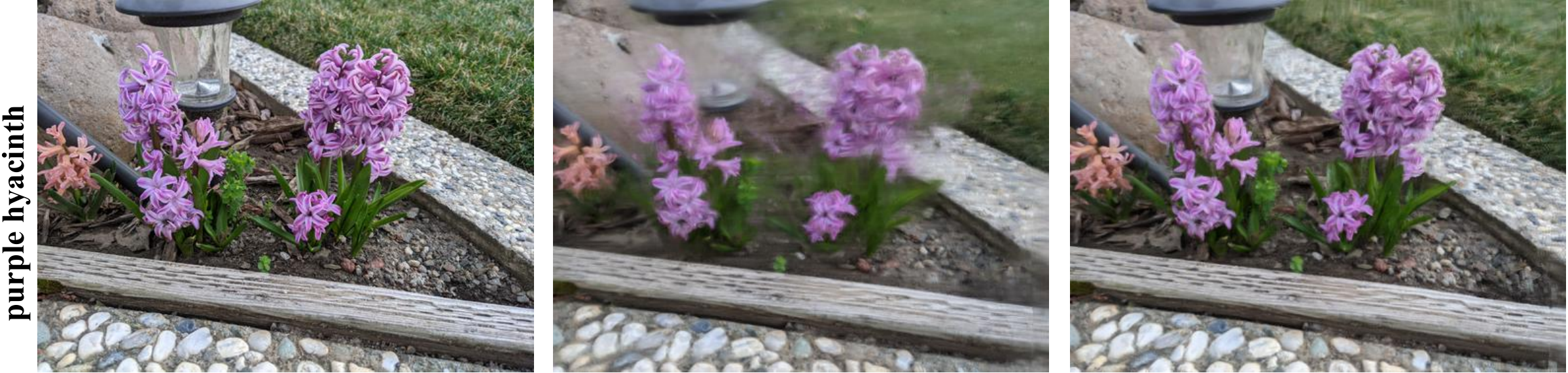}
        \label{fig:ibrnet_collected_purple_hyacinth}
    } \\
    % \subfloat {
    %     \includegraphics[width=0.95\linewidth]{ibrnet_collected_red_corvette}
    %     \label{fig:ibrnet_collected_red_corvette}
    % } \\
    \subfloat {
        \includegraphics[width=0.95\linewidth]{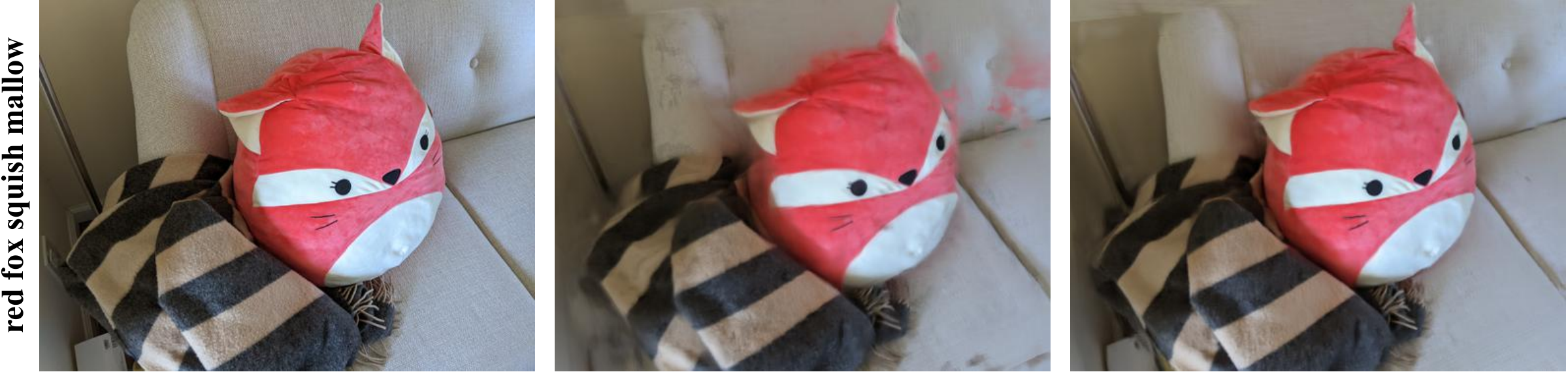}
        \label{fig:ibrnet_collected_red_fox_squish_mallow}
    } \\
    \subfloat {
        \includegraphics[width=0.95\linewidth]{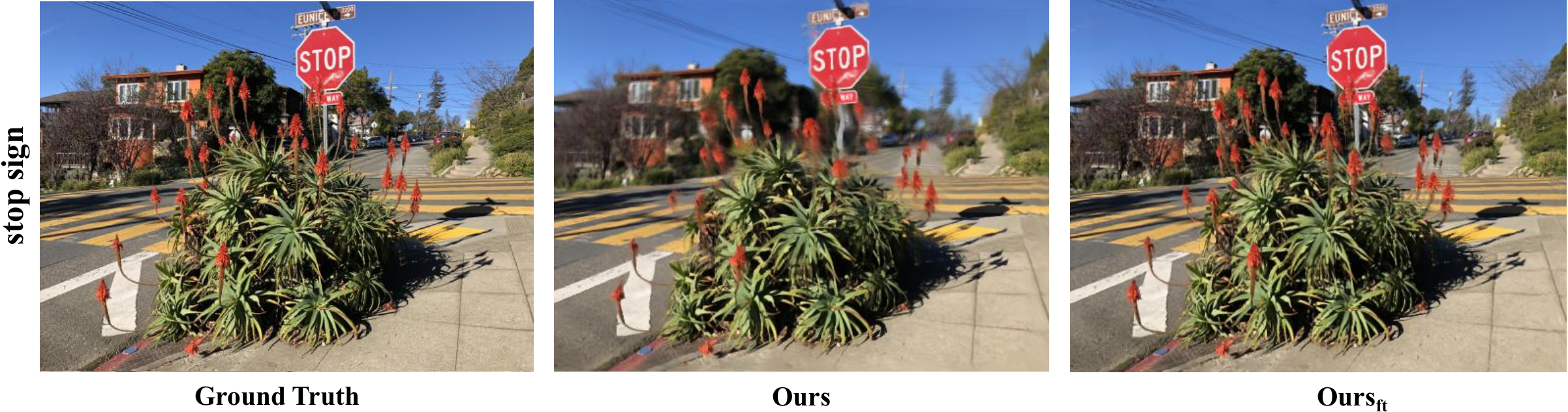}
        \label{fig:ibrnet_collected_stop_sign}
    }
    \caption{\textbf{The qualitative results on ibrnet self-collected dataset~\cite{DBLP:conf/cvpr/WangWGSZBMSF21}.}}
    \label{fig:ibrnet_collected_render}
\end{figure*}

\begin{figure*}[htbp]
    \centering
    \subfloat {
        \includegraphics[width=0.95\linewidth]{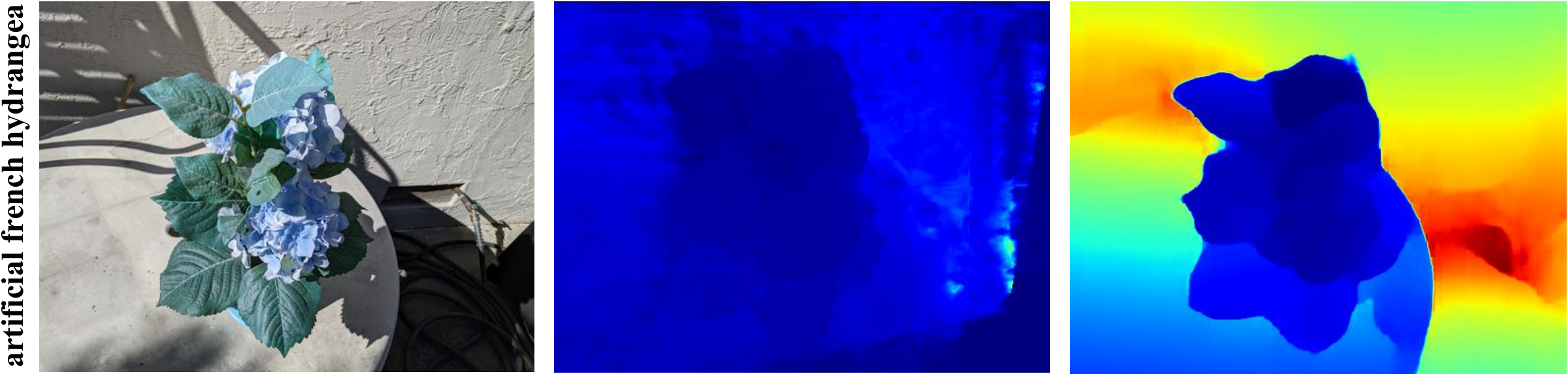}
        \label{fig:ibrnet_collected_artificial_french_hydrangea_depth}
    } \\
    \subfloat {
        \includegraphics[width=0.95\linewidth]{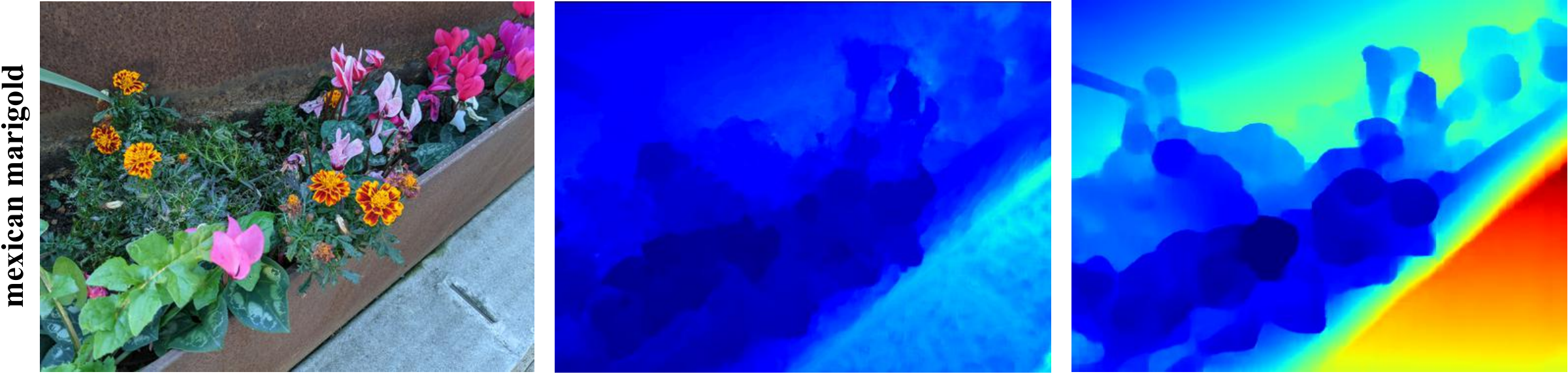}
        \label{fig:ibrnet_collected_mexican_marigold_depth}
    } \\
    \subfloat {
        \includegraphics[width=0.95\linewidth]{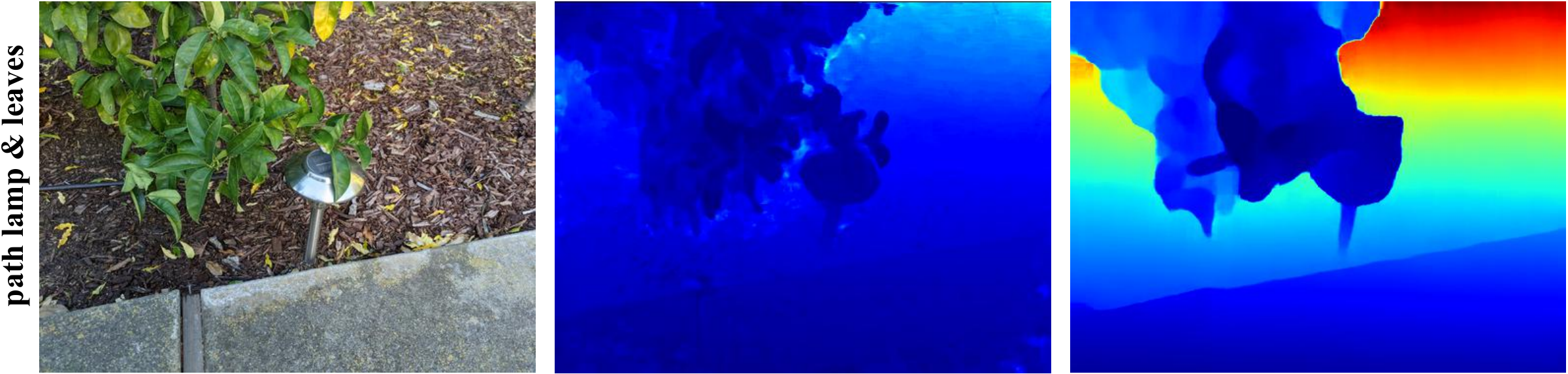}
        \label{fig:ibrnet_collected_path_lamp_leaves_depth}
    } \\
    \subfloat {
        \includegraphics[width=0.95\linewidth]{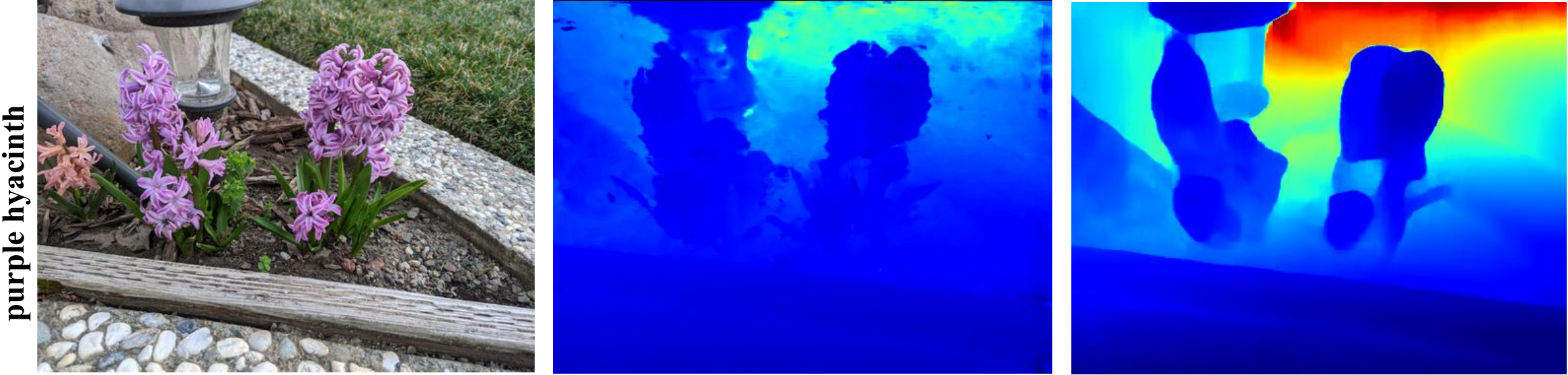}
        \label{fig:ibrnet_collected_purple_hyacinth_depth}
    } \\
    % \subfloat {
    %     \includegraphics[width=0.95\linewidth]{ibrnet_collected_red_corvette_depth}
    %     \label{fig:ibrnet_collected_red_corvette_depth}
    % } \\
    \subfloat {
        \includegraphics[width=0.95\linewidth]{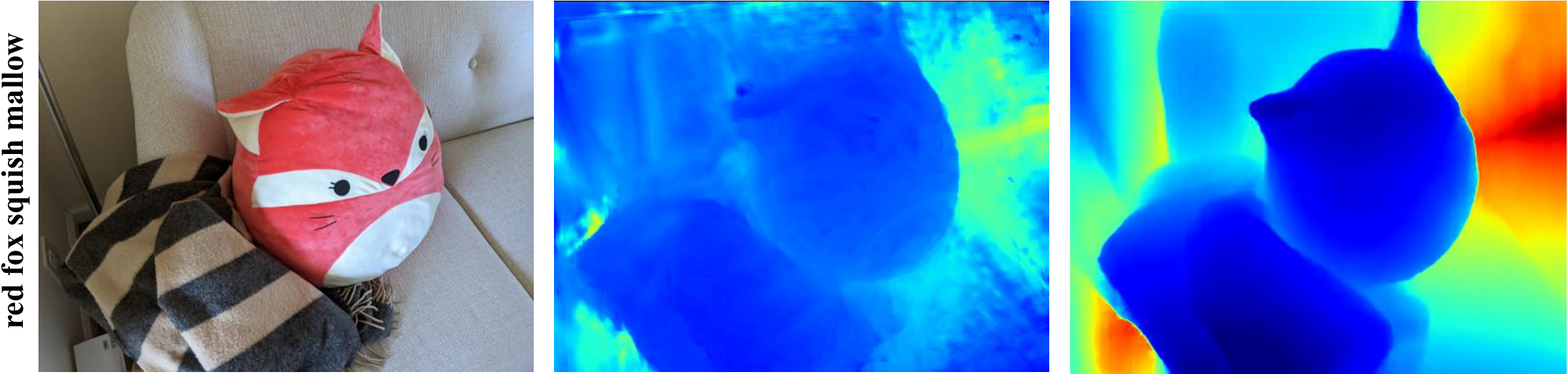}
        \label{fig:ibrnet_collected_red_fox_squish_mallow_depth}
    } \\
    \subfloat {
        \includegraphics[width=0.95\linewidth]{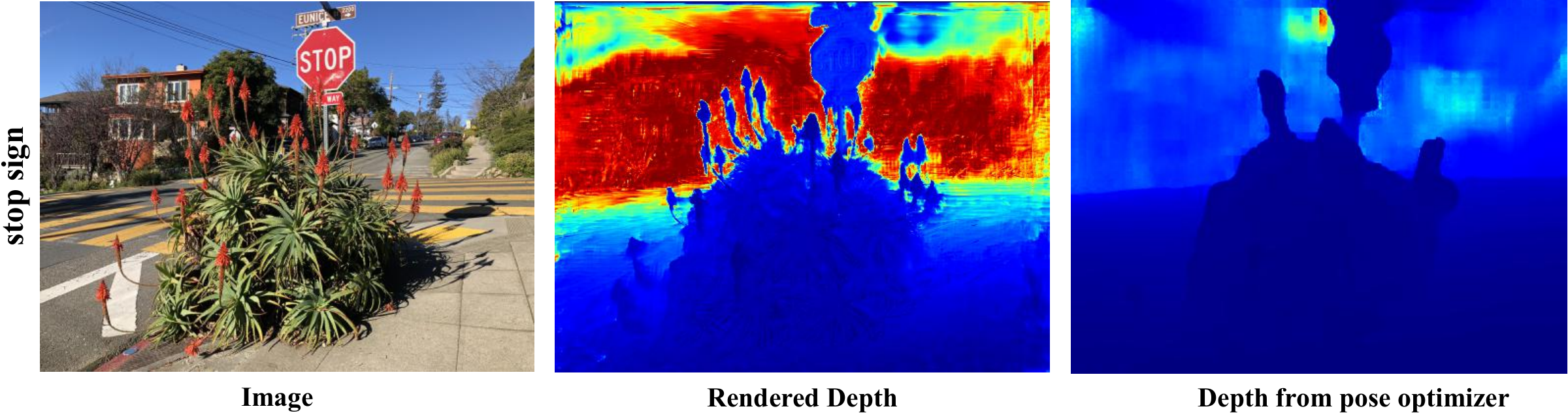}
        \label{fig:ibrnet_collected_stop_sign_depth}
    }
    \caption{\textbf{Depth visualization on ibrnet self-collected dataset~\cite{DBLP:conf/cvpr/WangWGSZBMSF21}.}}
    \label{fig:ibrnet_collected_render_depth}
\end{figure*}

To further support the effectiveness of our method, we evaluate our method on more IBRNet's self-collected 
dataset~\cite{DBLP:conf/cvpr/WangWGSZBMSF21}. The qualitative results of the rendered image and the corresponding geometry after finetuning are respectively given in Fig.~\ref{fig:ibrnet_collected_render} and Fig.~\ref{fig:ibrnet_collected_render_depth}. The quantitative results are given in Table~\ref{table:ibrnet_collected_quantitative_nvs}. We notice there is a drop in PSNR for `usps van'. This is because a large portion of the images in `usps van' contains the shadow of a moving human (see Fig.~\ref{fig:usps_van}), and our method cannot handle moving objects.

% ------------------------------------------ Table ---------------------------------------------------
\begin{table}[htbp]
    \centering
    \resizebox{0.5\textwidth}{!}{
      \begin{tabular}{l | c c | c c | c c }
        \toprule
  
        \multirow{3}{*}{Scenes}  &
        \multicolumn{2}{c|}{\textbf{PSNR}\ $\uparrow$} &
        \multicolumn{2}{c|}{\textbf{SSIM}\ $\uparrow$} &  
        \multicolumn{2}{c}{\textbf{LPIPS}\ $\downarrow$} \\
        
        \cmidrule(r){2-3} \cmidrule(r){4-5} \cmidrule(r){6-7}
        & $\text{Ours}$ & $\text{Ours}_{\text{ft}}$
        
        & $\text{Ours}$ & $\text{Ours}_{\text{ft}}$
        
        & $\text{Ours}$ & $\text{Ours}_{\text{ft}}$ \\
        
    \midrule
  
        red corvette & 18.04 & 19.46 & 0.728 & 0.785 & 0.345 & 0.238 \\
        
        usps van & 16.96 & 16.68 & 0.761 & 0.766 & 0.258 & 0.208 \\

        path lamp \& leaves & 19.10 & 20.34 & 0.403 & 0.56 & 0.497 & 0.275 \\

        purple hyacinth & 18.19 & 19.48 & 0.387 & 0.506 & 0.448 & 0.269 \\

        artificial french hydrangea & 17.61 & 19.03 & 0.531 & 0.600 & 0.470 & 0.292 \\

        red fox squish mallow & 23.26 & 24.37 & 0.668 & 0.700 & 0.358 & 0.270 \\

        mexican marigold & 20.59 & 21.48 & 0.506 & 0.582 & 0.419 & 0.240 \\

        stop sign & 21.27 & 22.04 & 0.738 & 0.803 & 0.195 & 0.099 \\

        \bottomrule
      \end{tabular}
    }
    \vspace{-2mm}
    \caption{Quantitative results of novel view synthesis on ibrnet self-collected dataset~\cite{DBLP:conf/cvpr/WangWGSZBMSF21}. 
             $\text{Ours}_{\text{ft}}$ denotes our results after finetuned $60,000$ iterations.}
    \vspace{-4mm}
    \label{table:ibrnet_collected_quantitative_nvs}
\end{table}

%%%%%%%%%%%%%%%%%%%%%%%%%%%%%%%%%%%%%%%%%%%%%%%%%%%% Results on Scannet %%%%%%%%%%%%%%%%%%%%%%%%%%%%%%%%%%%%%%%%%%%%%
\subsection{Results on ScanNet Datasets}

\begin{figure*}[htbp]
    \centering
    \subfloat {
        \includegraphics[width=0.95\linewidth]{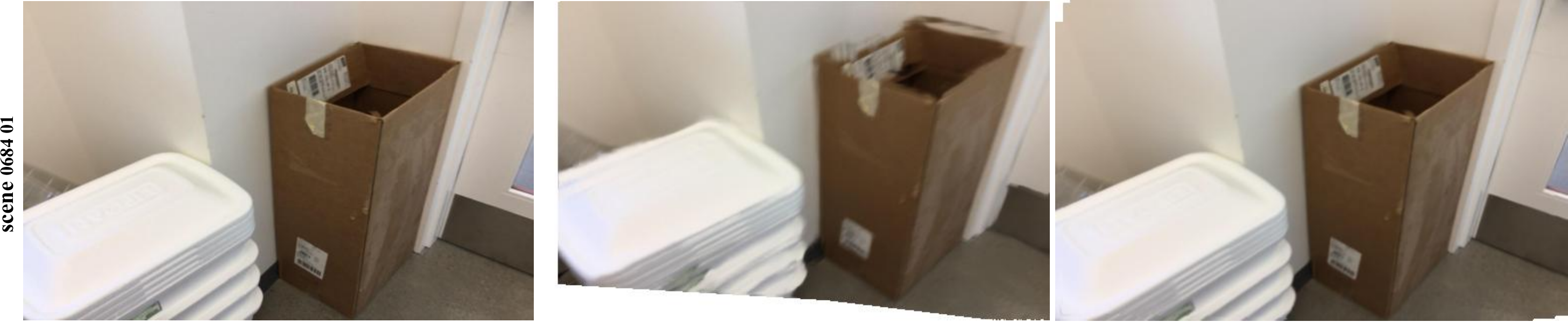}
        \label{fig:scannet_scene0684_01}
    } \\
    \subfloat {
        \includegraphics[width=0.95\linewidth]{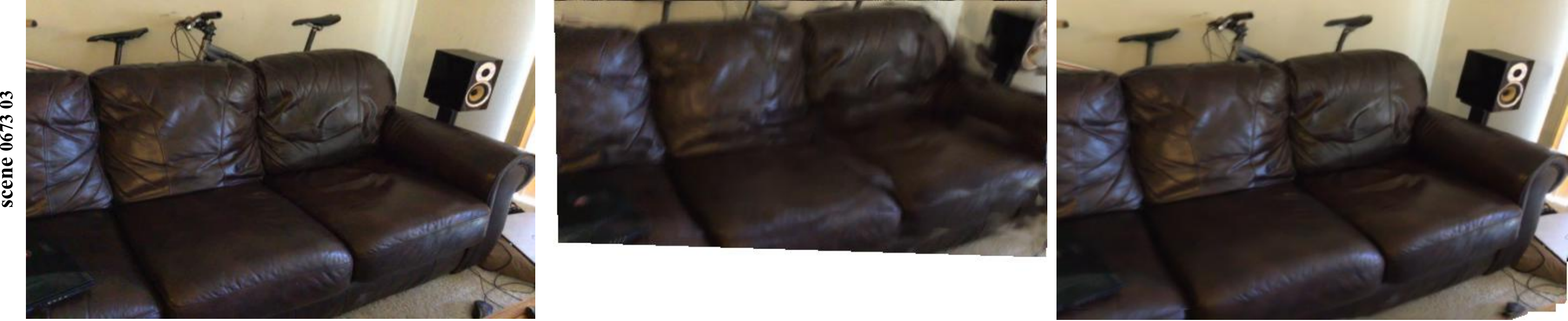}
        \label{fig:scannet_scene0673_03}
    } \\
    \subfloat {
        \includegraphics[width=0.95\linewidth]{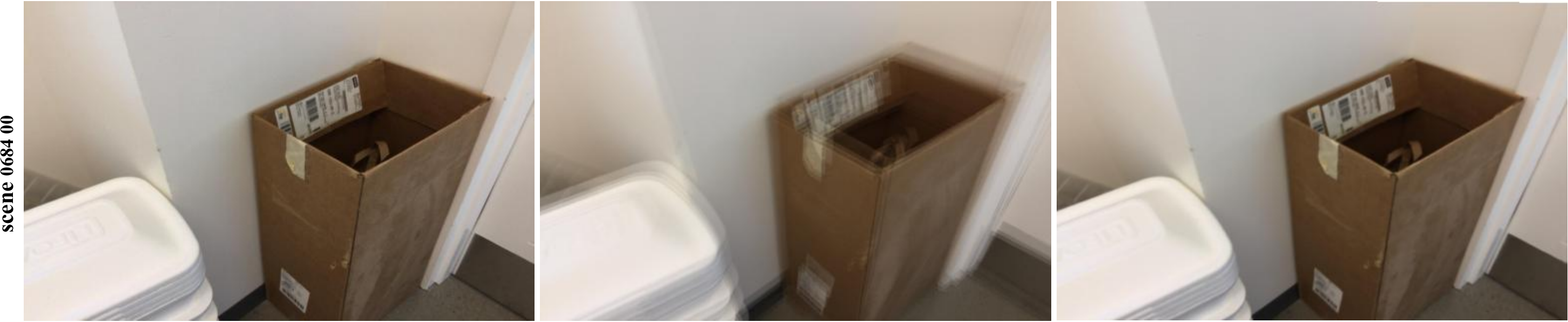}
        \label{fig:scannet_scene0684_00}
    } \\
    \subfloat {
        \includegraphics[width=0.95\linewidth]{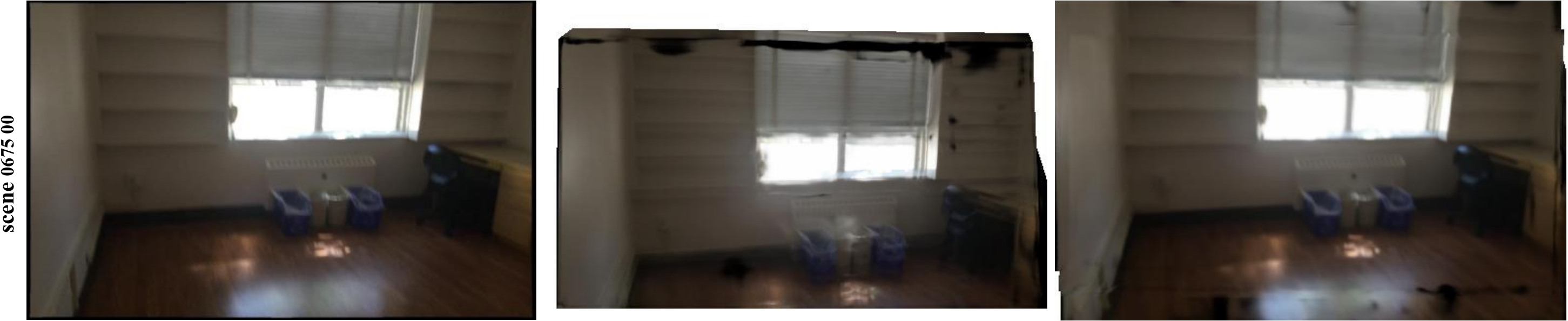}
        \label{fig:scannet_scene0675_00}
    } \\
    \subfloat {
        \includegraphics[width=0.95\linewidth]{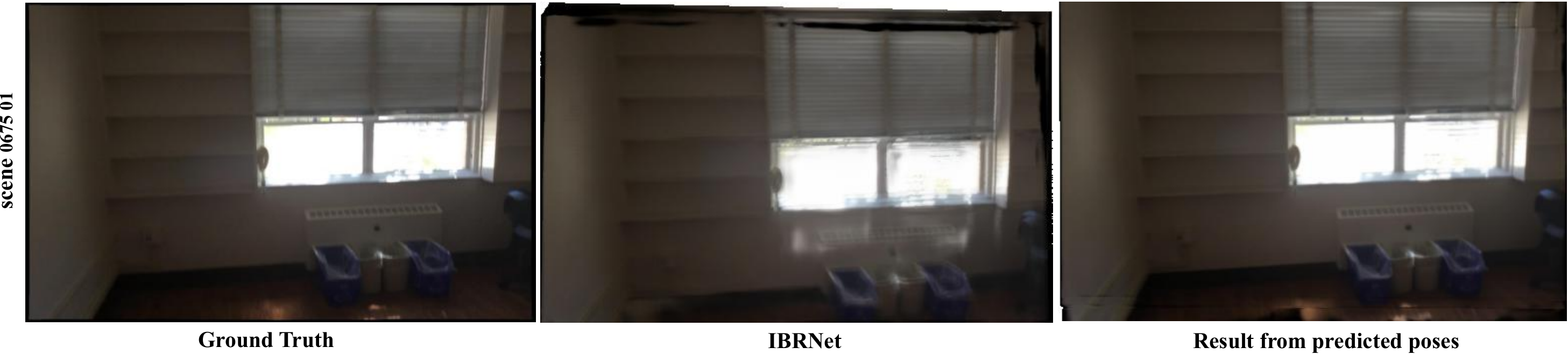}
        \label{fig:scannet_scene0675_01}
    }
    \caption{\textbf{The qualitative results on ScanNet dataset~\cite{DBLP:conf/cvpr/DaiCSHFN17}.}}
    \label{fig:scannet_render}
\end{figure*}

\begin{figure*}[htbp]
    \centering
    \subfloat {
        \includegraphics[width=0.95\linewidth]{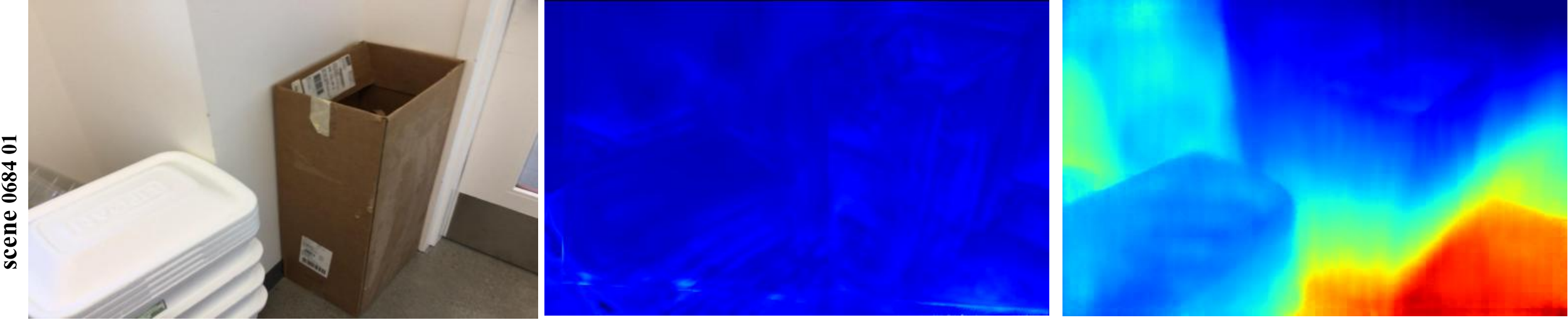}
        \label{fig:scannet_depth3}
    } \\
    \subfloat {
        \includegraphics[width=0.95\linewidth]{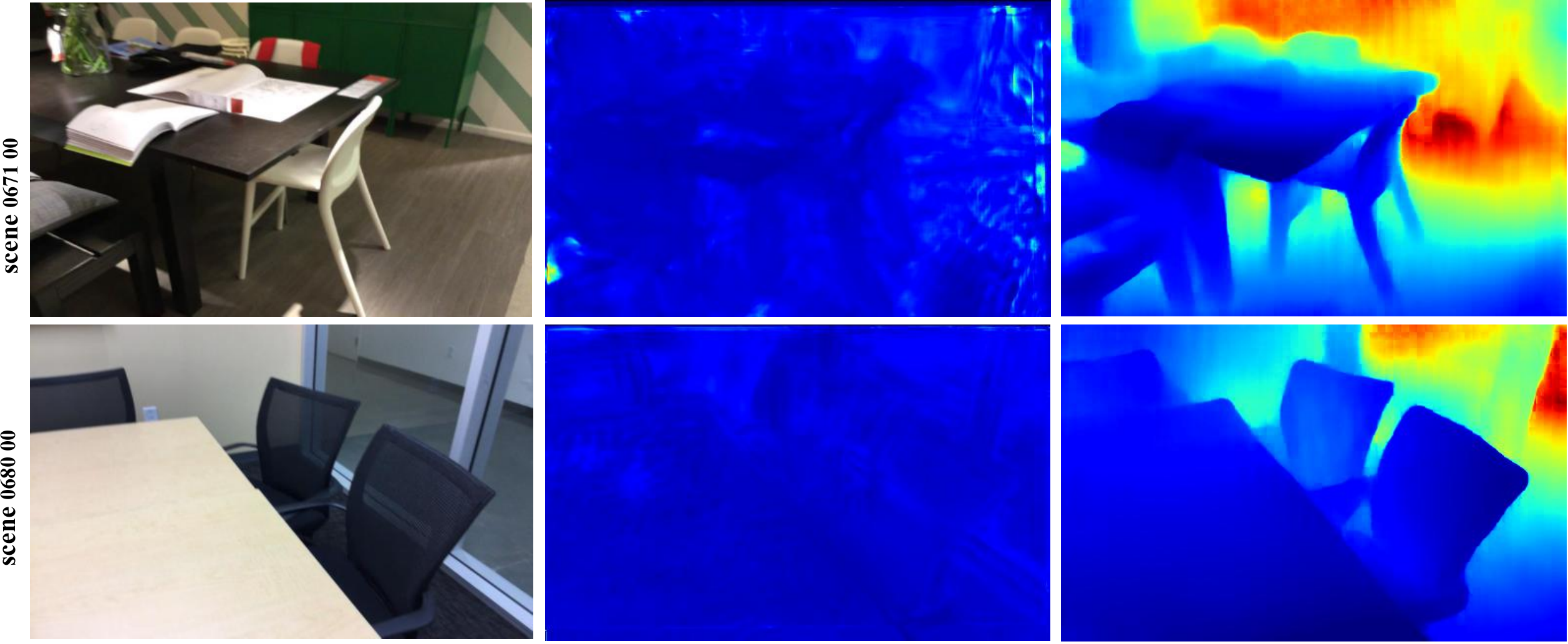}
        \label{fig:scannet_depth1}
    } \\
    \subfloat {
        \includegraphics[width=0.95\linewidth]{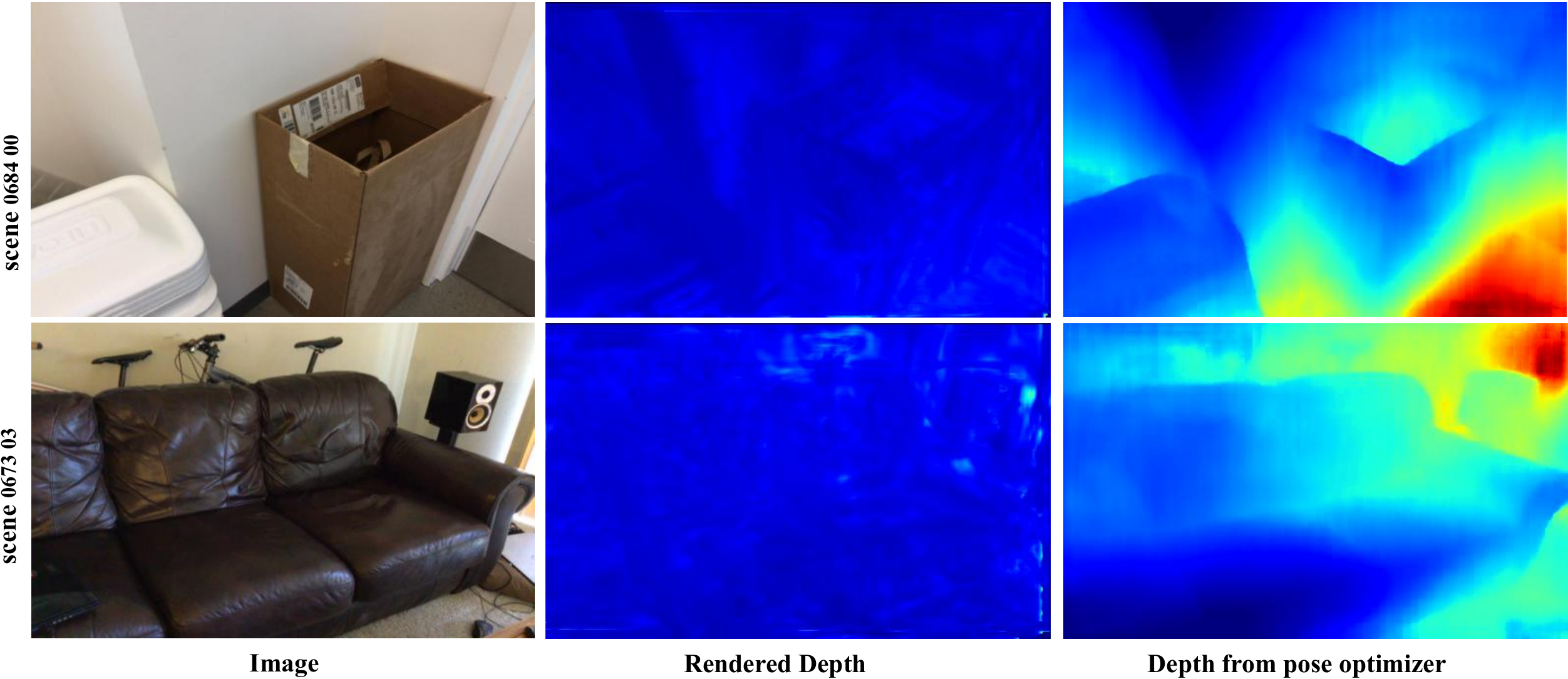}
        \label{fig:scannet_depth2}
    }
    \caption{\textbf{Depth visualization on ScanNet dataset~\cite{DBLP:conf/cvpr/DaiCSHFN17}.}}
    \label{fig:scannet_render_depth}
\end{figure*}

We also show the results evaluated on 7 scenes of the ScanNet dataset~\cite{DBLP:conf/cvpr/DaiCSHFN17}. 
The qualitative results of the rendered image and the corresponding geometry after finetuning 
are respectively given in Fig.~\ref{fig:scannet_render} and Fig.~\ref{fig:scannet_render_depth}. The quantitative results 
are given in Table~\ref{table:scannet_quantitative_nvs}. We can observe that our method outperforms IBRNet by a large margin. The poor 
performance of IBRNet on this dataset is due to the inaccurate camera poses. However, our method does not rely on precomputed camera poses 
and the regressed camera poses are accurate to enable high-quality image rendering.

% ------------------------------------------ Table ---------------------------------------------------
\begin{table}[htbp]
    \centering
    \resizebox{0.5\textwidth}{!}{
      \begin{tabular}{c | c c | c c | c c }
        \toprule
  
        \multirow{3}{*}{Scenes}  &
        \multicolumn{2}{c|}{\textbf{PSNR}\ $\uparrow$} &
        \multicolumn{2}{c|}{\textbf{SSIM}\ $\uparrow$} &  
        \multicolumn{2}{c}{\textbf{LPIPS}\ $\downarrow$} \\
        
        \cmidrule(r){2-3} \cmidrule(r){4-5} \cmidrule(r){6-7}
        & IBRNet~\cite{DBLP:conf/cvpr/WangWGSZBMSF21} & Ours
        
        & IBRNet~\cite{DBLP:conf/cvpr/WangWGSZBMSF21} & Ours
        
        & IBRNet~\cite{DBLP:conf/cvpr/WangWGSZBMSF21} & Ours \\
        
    \midrule
  
        scene0671-00 & 12.29 & 26.60 & 0.518 & 0.910 & 0.451 & 0.113 \\
        
        scene0673-03 & 11.31 & 23.56 & 0.457 & 0.859 & 0.615 & 0.156 \\

        scene0675-00 & 10.55 & 19.95 & 0.590 & 0.875 & 0.589 & 0.207 \\

        scene0680-00 & 14.69 & 31.05 & 0.709 & 0.958 & 0.389 & 0.056 \\

        scene0684-00 & 18.46 & 33.61 & 0.737 & 0.975 & 0.296 & 0.052 \\

        scene0675-01 & 10.33 & 23.56 & 0.595 & 0.899 & 0.548 & 0.166 \\

        scene0684-01 & 14.69 & 33.01 & 0.678 & 0.967 & 0.426 & 0.056 \\ 

        \bottomrule
      \end{tabular}
    }
    \vspace{-2mm}
    \caption{Quantitative results of novel view synthesis on ScanNet dataset~\cite{DBLP:conf/cvpr/DaiCSHFN17} after finetuning.}
    \vspace{-4mm}
    \label{table:scannet_quantitative_nvs}
\end{table}

\newpage
\clearpage
%%%%%%%%% REFERENCES
{\small
\bibliographystyle{ieee_fullname}
\bibliography{egbib}
}

\end{document}